%% file: eccv2016submission.tex
\documentclass[runningheads]{llncs}
\usepackage{graphicx}
\usepackage{amsmath,amssymb} 
\usepackage{color}
\usepackage{comment}
\usepackage{epstopdf} 
\usepackage[table,usenames,dvipsnames]{xcolor}%
\usepackage[width=122mm,left=12mm,paperwidth=146mm,height=193mm,top=12mm,paperheight=217mm]{geometry}
\usepackage{color, colortbl}
\usepackage{hhline}
\usepackage{multirow}%

\begin{document}
\pagestyle{headings}
\mainmatter
\def\ECCV18SubNumber{2834}  

\definecolor{dgreen}{rgb}{0.0,0.6,0.0} 
\definecolor{dred}{rgb}{0.6,0.0,0.0} 
\definecolor{BrickRed}{rgb}{0.72,0.0,0.0}%
\definecolor{grey}{rgb}{0.6,0.6,0.6}%
\newcommand{\tce}[1]{\textcolor{orange}{#1}}	      
\newcommand{\tcb}[1]{\textcolor{dred}{\textbf{#1}}}	  
\newcommand{\tco}[1]{\textcolor{cyan}{#1}}	          
\newcommand{\todo}{$\clubsuit$} 
\newcommand{\TODO}{$\clubsuit$\textbf{TODO}$\clubsuit$}

\title{FusionNet and AugmentedFlowNet: \\ Selective Proxy Ground Truth \\for Training on Unlabeled Images} 

\author{
Osama Makansi\textsuperscript{*} \and 
Eddy Ilg\textsuperscript{*} \and 
Thomas Brox}
\institute{
University of Freiburg, Germany \\
\email{\{makansi,ilg,brox\}@cs.uni-freiburg.de}
}

\authorrunning{O. Makansi, E. Ilg and T. Brox}

\titlerunning{FusionNet and AugmentedFlowNet}

\maketitle

\newcommand{\pz}{\phantom{0}}

\begin{abstract}
\vspace*{-4mm}
Recent work has shown that convolutional neural networks (CNNs) can be used to estimate optical flow with high quality and fast runtime. This makes them preferable for real-world applications. However, such networks require very large training datasets. Engineering the training data is difficult and/or laborious. This paper shows how to augment a network trained on an existing synthetic dataset with large amounts of additional unlabelled data. In particular, we introduce a selection mechanism to assemble from multiple estimates a joint optical flow field, which outperforms that of all input methods. The latter can be used as proxy-ground-truth to train a network on real-world data and to adapt it to specific domains of interest. Our experimental results show that the performance of networks improves considerably, both, in cross-domain and in domain-specific scenarios. As a consequence, we obtain state-of-the-art results on the KITTI benchmarks. 
\end{abstract}

\section{Introduction}
{
\renewcommand{\thefootnote}{\fnsymbol{footnote}}
\footnotetext[1]{equal contribution}
}
\vspace*{-2mm}
Like all deep learning applications that follow the supervised learning paradigm, the success of learning optical flow estimation stands and falls with the availability and quality of training data. In case of optical flow, the creation of ground-truth annotation on real images is extremely tedious and virtually impossible on large datasets. 
For this reason, state-of-the-art networks for optical flow estimation, such as FlowNet 2.0~\cite{flownet2} and PWC-Net~\cite{PWCnet} have been trained on synthetically rendered images.
These networks tend to generalize comparatively well to real images -- in contrast to semantic tasks, such as object detection or semantic segmentation. This is because correspondence estimation is different from recognition and does not depend so much on the content of the images. In fact, optical flow estimation is possible without any learning, thus not requiring any training data. There is a long history of unsupervised optical flow methods that implement the concept of correspondence. These classical methods perform equally well as a state-of-the-art optical flow network, yet with significantly higher runtimes.

The advantage of learning comes in when correspondences cannot be established easily and priors are needed to make decisions. Typical examples are areas in the image that have homogeneous color (aperture problem) or areas that are occluded in the other image. Works from the pre-learning era used handcrafted regularizers~\cite{schunck,meminperez} and corresponding optimization heuristics to hallucinate optical flow in these areas. Learning such priors is much more elegant and also more successful: networks tend to outperform these classical techniques especially in occluded areas. However, such learning of priors is no longer independent of the image content: while basic hallucination strategies for occluded regions can be estimated from synthetic data, the hallucinated content should ideally depend on the objects in the scene. Thus, there is a domain gap between synthesized training images and real images, just like in semantic tasks. Real images are required for training.
Multiple strategies have been proposed to integrate real images into the training procedure. These span from using the same unsupervised training loss for the network as is used in variational methods~\cite{ahmadi,unflow}, over multi-task learning with an auxiliary task that allows learning from unlabelled images~\cite{SZB17}, to training on pseudo-ground-truth obtained from running an (unsupervised) variational method~\cite{guided_flow_17}.

This paper comprises two contributions. First, we present an assessment network that learns to predict the error for each of a set of flow fields generated with various optical flow estimation techniques. Then, a fused optical flow field can be trivially obtained by selecting for each pixel the flow vector with the smallest predicted error. We show that this assessment network, which we call FusionNet combines the advantages of a potentially large set of techniques and avoids their limitations. As a consequence, FusionNet yields results that exceed the performance of all methods that produced its input. Independent on how the state of the art will improve in the future, FusionNet can always benefit from these improvements. However, this comes at the cost of very large runtimes, since a whole set of partially slow methods must be run on the test image for the assessment network to assemble the final flow field. This is a show stopper for most optical flow applications.

Thus, as a second contribution, we augment a FlowNet by training it on the flow obtained with the assessment network, which now serves as proxy ground-truth. This shifts the large runtimes to the training phase, while the final network is as fast as a regular FlowNet at test time. Training data can be generated on all sorts of unlabeled videos, which allows the augmented FlowNet to learn priors from real images. This yields the currently best accuracy-runtime trade-off and enables the specialization to target domains directly on real images without tedious modeling of synthetic scenes in such domains. We show-case this with state-of-the-art results on the KITTI benchmarks. 


\section{Related Work}

\textbf{Traditional optical flow estimation.} 
Optical flow estimation goes back to the works of Lucas\&Kanade~\cite{lucaskanade} and Horn\&Schunck~\cite{schunck}. Both rely on a brightness constancy term combined with a
local or global smoothness assumption. Especially the variational approach of Horn\&Schunck was extended by many successive works~\cite{meminperez,Bro04a,pocktvl1}. 
While variational methods are very precise in small displacement cases, they have deficits in case of large displacements. This was taken into account by Brox et al.~\cite{ldof}, who mixed the variational method with a simple nearest-neighbor matching of local descriptors. DeepMatching~\cite{deepmatching} elaborated on the matching, and EpicFlow~\cite{epicflow} improved the variational refinement. 
FlowFields~\cite{flowfields} builds upon EpicFlow and elaborates on the matching using a random search strategy. 
The present state of the art is defined by DCFlow~\cite{dcflow} and MRFlow~\cite{Wulff:CVPR:2017}.
The accuracy of these techniques is very high and on-par or even higher than with learning based techniques. However, the combinatorial search in state-of-the-art methods leads to quite large runtimes that do not allow for interactive frame rates. 


\textbf{Optical flow with supervised learning.}
End-to-end learning of optical flow was pioneered by the work of Dosovitskiy et al.~\cite{flownet}, which presented the two network architectures FlowNetS and FlowNetC. The former is purely convolutional, while the latter includes an explicit correlation. The networks were trained on a simplistic dataset made from Flickr and chair images to which affine transformations were applied (FlyingChairs). Mayer et al.~\cite{dispnet} introduced a more sophisticated 3D dataset (FlyingThings3D). 
Ilg et al.~\cite{flownet2} presented a stack of networks termed FlowNet 2.0 with high accuracy and fast runtime.
Ranjan et al.~\cite{spynet} presented a network architecture that contains a spatial pyramid and runs even faster than FlowNet 2.0, but at the cost of accuracy. Sun et al.~\cite{PWCnet} extended this idea by introducing correlations at the different pyramid levels. Their network termed PWC-Net currently achieves state-of-the-art results. 

Other methods combine feature learning with traditional methods: FlowFieldsCNN~\cite{FlowFieldsCNN} uses an improved hinge embedding loss to train a Siamese architecture for feature extraction, which is then used in combination with FlowFields. PatchBatch~\cite{PatchBatch} shows that CNN features can even be improved to a level on which plain nearest-neighbor matching performs well. DeepDiscreteFlow~\cite{DeepDiscreteFlow} combines a local network with a context network and discrete optimization. 

\textbf{Optical flow with unsupervised learning components.} 
Ahmadi et al.~\cite{ahmadi} proposed an unsupervised learning approach by using the brightness constancy loss from variational approaches to train a CNN. In principle, their approach replaces the Gauss-Newton step in variational optimization with back-propagation on a network representation. While coming from a fully unsupervised approach, the resulting flow fields are inferior to those of unsupervised variational techniques. Meister et al.~\cite{unflow} proposed an additional unsupervised loss based on forward-backward consistency to train the network termed UnFlow in a completely unsupervised manner. 

Several other methods introduce unsupervised losses in addition to supervised training on synthetic data. Yu et al.~\cite{backtobasics} and Ren et al.~\cite{dstflow} use the loss from variational approaches to refine the decoder stages of a pre-trained FlowNet. 
Lai et al.~\cite{NIPS2017_6639} use a GAN approach to distinguish the optical flow estimated by the generator network from ground-truth optical flow.   
Sedaghat et al. \cite{SZB17} proposed the self-supervised auxiliary task of next frame prediction as additional loss. Like the above-mentioned works, this allows them to improve FlowNet on real-world data. 
The guided optical flow proposed by Zhu et al.~\cite{guided_flow_17} uses the flow computed by a traditional, unsupervised method as proxy ground-truth to train a network in the usual supervised manner. 
The final network is limited by the performance of the traditional method that provided the proxy ground-truth. 
In contrast, we anticipate this drawback by training on flow fields produced by multiple different methods and locally selecting the best. This way, the final network can yield better results than any single method that produced the training data. 

\textbf{Optical flow fusion.}
The principle to locally select the best flow vector from a set of flow fields has been implemented outside the scope of deep learning. Lempitsky et al.~\cite{lempitskyCVPR08} proposed a combinatorial optimization approach to combine flow fields from multiple methods based on a smoothness loss. Their approach was also used in MDPFlow~\cite{mdpflow}, which locally combined multiple hypotheses from coarser pyramid levels and nearest neighbor matching.
In contrast to these approaches, our selection among flow vectors is based on a deep network that learns to predict directly the optical flow error rather than just selecting based on smoothness priors. 


\section{FusionNet}

\begin{figure}[t]
    \begin{center} 
        \includegraphics[width=0.75\textwidth]{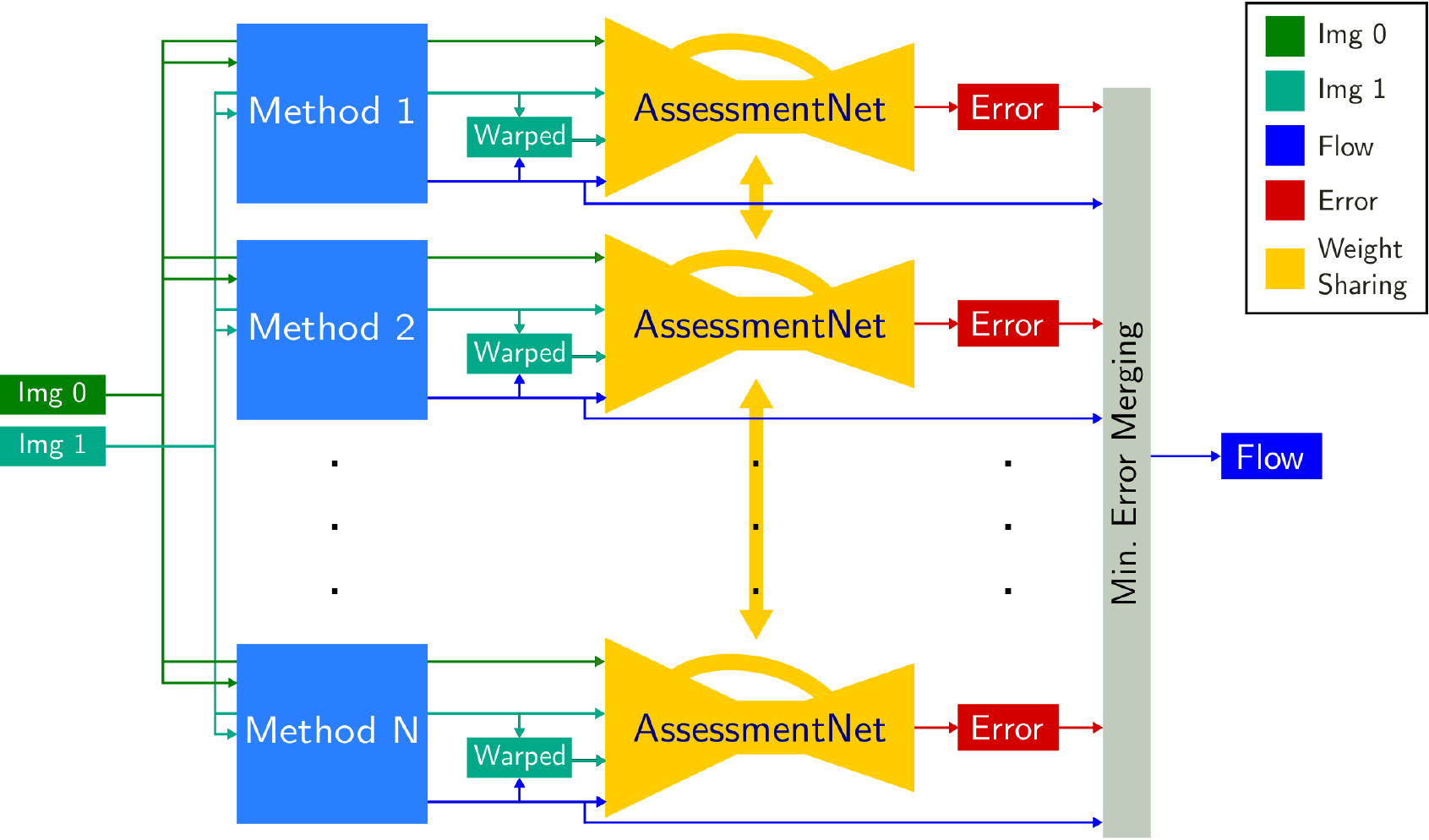}
    \end{center} 
    \caption{        
        Overview of the FusionNet principle. Given the input images, the optical flow is estimated with various existing methods. Each method's optical flow estimate is used to warp the second image. The two input images, the warped image, and the flow are fed into the proposed assessment network, which is trained on predicting the error of each flow field. Finally the flow fields are merged by locally choosing the flow vector with the minimum predicted error. 
        \label{fig:fusion_net}
    } 
\end{figure}

We assume that various optical flow estimation methods have different strengths and weaknesses. This does not exclude that these methods may have also many difficulties in common. However, as long as there are differences, we want to exploit these differences to choose from the method that works best on a particular problem. 

To this end, we propose an assessment network that predicts the errors of the optical flow estimated by a set of existing methods, as shown in Figure~\ref{fig:fusion_net} and is trained on synthetic data with available ground-truth optical flow. On first glance, this training on synthetic images looks like we are back at square one. However, the task of assessment is different from the task of flow estimation itself. First, we benefit from the information contained in the various input flow fields. Second, the assessment task may generalize more easily to other domains than the task of optical flow estimation, since it must only find ways to predict errors rather than predicting the flow field itself. 

The assessment network uses a typical encoder-decoder architecture with skip-connections; the architecture details are as in FlowNetS~\cite{flownet2}. It takes the two input images into account together with the flow estimate and the second image warped by that flow. 
This error map predicted by the assessment network is used to optimally combine the estimated flow fields. We refer to the complete setup, as shown in Figure~\ref{fig:fusion_net}, as FusionNet. We investigate two different loss functions: an L1 loss and a hinge loss. 

\subsection{L1 Loss}
For training the assessment network with an L1 loss, we let the network directly estimate the pixel-wise endpoint error. Let the estimated optical flow at a certain pixel be denoted $w=(u,v)$ with the x- and the y-components $u$ and $v$. The ground-truth endpoint error $e_\mathrm{gt}$ for a pixel location is: 
\begin{equation}
e_\mathrm{gt} = \sqrt{(u - u_{\mathrm{gt}})^2+(v - v_{\mathrm{gt}})^2}.
\end{equation} 
Let $e$ be the error predicted by the assessment network. To improve on the predicted error, we apply back-propagation with the L1 loss: 
\begin{equation}
\mathcal{L}_1(e) = |e_\mathrm{gt} - e|.
\end{equation} 
In principle, one could train a separate, specialized network for each input method to be assessed. However, since we want to improve on the generalization of the assessment network, we use the same network for assessing all input flows, and rather sample the mini-batches during training from the different methods. More training details are provided in Section~\ref{training_details}.

\subsection{Hinge Loss}

%
%
%

Directly applying an L1 loss on the error makes the network estimate the error for each method. However, for the fusion we only need to know the input methods with the lowest error. That means, the L1 loss potentially solves a harder problem than necessary to reach the actual goal\footnote{That said, the prediction of the error could be valuable on its own right for a series of other purposes not discussed in this paper, for instance, uncertainty estimation.}.
A related problem to picking the input with the smallest error is the one of designing a distance metric to match patches. 
This metric only needs to reflect the ranking, e.g. "A is closer to B than A is to C"~\cite{NIPS2003_2366,weinberger2009distance}. 
Many feature learning algorithms use this as a triplet loss~\cite{dcflow,wang01,tripletnet,fastnet,wohlhart01}. 
With the same motivation, we use the well-known multi-class hinge loss~\cite{multihinge,multisvm}
\begin{equation}
    \label{joint_ass_function}
         	\mathcal{L}_{\rm Margin}(e_1,...,e_N) = \sum_{i\neq j} \max(0,m+e_j-e_i),
\end{equation}

where $j$ is the index of the method with the lowest error according to the ground-truth,
and $m$ is the minimum margin between the best estimate and the other estimates. If the predicted best error corresponds to the true index $j$ and all other errors are at least $m$ larger than $e_j$, this loss will be zero. Otherwise, each error that is above the allowed margin will contribute to the loss.
Since the network is allowed to rescale the errors, we can set $m=1$ without loss of generality. Note that the errors predicted by the network with this loss do no longer correspond to the L1 error but may be rescaled. The rescaling factor may even be different for each pixel.  
Obviously, the hinge loss implies joint training of the assessment network while giving all $N$ methods as input. 



\begin{figure}[t]
    \begin{center} 
        \includegraphics[width=0.35\textwidth]{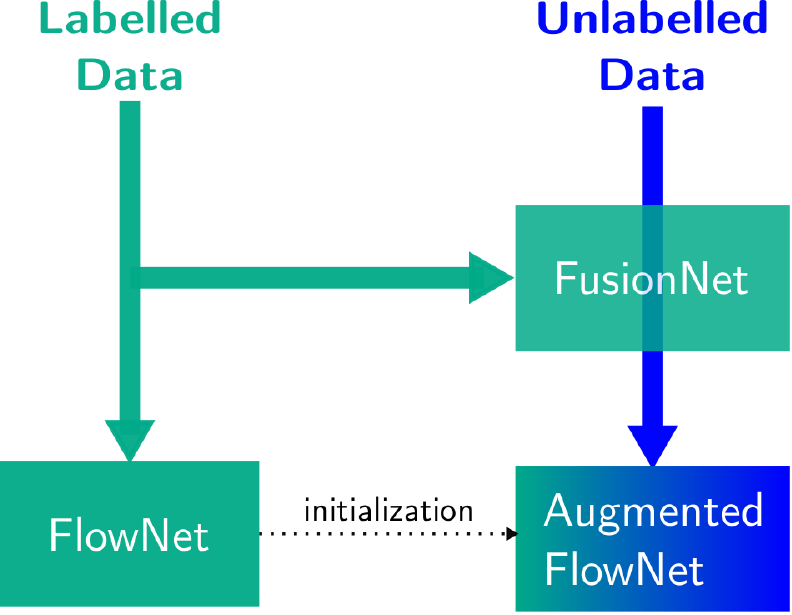}
    \end{center} 
    \caption{        
        Using our FusionNet to augment a FlowNet: FlowNet and FusionNet are trained on labeled data. Subsequently, FusionNet is used to augment FlowNet with large amounts of unlabeled data. 
        \label{fig:domain_transfer}
	\vspace*{-3mm} 
    } 
\end{figure}

\section{Augmented FlowNet}

Given the FusionNet from the last section, we can apply it to any unlabelled data to estimate high-quality optical flow. However, running FusionNet is very costly, since it requires running the various, partially very slow optical flow estimation methods. In order to have fast optical flow estimation at test time, we use the optical flow fields estimated with FusionNet as proxy ground-truth in order to finetune a FlowNet, for instance, to optimize it for a specific domain or to make it run better on general real-world videos. The principle is quite straightforward  and illustrated in Figure~\ref{fig:domain_transfer}. 


\section{Experiments}

We evaluated the concept on the common optical flow benchmarks, where we can quantify the improvements by the fusion and by the augmentation of FlowNet directly. 
In addition, we demonstrate the effect of the augmentation in a motion segmentation context.


\subsection{Training Details} \label{training_details}
For training the assessment network, we followed the same training schedule as proposed in Ilg et al.~\cite{flownet2} for training FlowNet, i.e., we first train on FlyingChairs for 1.2m iterations and subsequently on FlyingThings3D for 500k iterations. 
The augmented FlowNet is initialized with a FlowNet trained on the same schedule. We also applied the same data augmentation mechanism, i.e., a set of spatial and color transformations. 
The networks were implemented using the Caffe framework. The code will be made publicly available upon publication. 

\subsection{Datasets}

We used the two publicly available synthetic datasets FlyingChairs\cite{flownet} and FlyingThings3D\cite{flownet2} to train the assessment network and the initial FlowNet before augmentation. These are the two datasets for which labeled training data is available. 
For the unsupervised fine-tuning, we use various unlabeled datasets that we grouped to two domains: animation movies and driving.  


\textbf{Animation movies.}
We collected several animation movies from the Blender project\cite{blender} and used them for unsupervised training. For such animation movies there is the potential option to derive ground-truth optical flow, as shown in Butler et al.~\cite{Butler:ECCV:2012} and Mayer et al.~\cite{MIFDB16}, but we did not use this option here and rather used just the unlabeled videos for training. 
For the evaluation in this domain we used the official Sintel benchmark dataset~\cite{Butler:ECCV:2012}.

\textbf{Driving.}
Driving scenes are a popular application domain for optical flow. Thus, we selected them to make a second evaluation domain. For unsupervised training, we took approximately 100k frames from the Frankfurt part of the publicly available Cityscapes dataset~\cite{Cordts2016Cityscapes}.  
For the evaluation in this domain, we used the two publicly available KITTI2012~\cite{Geiger2012CVPR} and KITTI2015~\cite{Menze2015CVPR} benchmark datasets. 

\textbf{Motion Segmentation.}
For indirect evaluation of the optical flow on a motion segmentation task, we used approximately 32k frames from the UdG-MS19 and UdG-MS20 datasets~\cite{udg} for unsupervised training.
We evaluated the motion segmentation on the FBMS benchmark dataset~\cite{OB14b}.


\subsection{FusionNet}

We evaluated the FusionNet with the following optical flow estimation techniques as input: LDOF~\cite{ldof}, DeepFlow~\cite{deepmatching}, EpicFlow~\cite{epicflow}, FlowFields~\cite{flowfields}, and FlowNet2~\cite{flownet2}. There are some very recent methods with even better performance, such as DCFlow~\cite{dcflow}, PWC-Net~\cite{PWCnet}, and MR-Flow\cite{Wulff:CVPR:2017}, but their code was not operational in time to include them for the experiments. A nice property of FusionNet is that new methods can be integrated trivially at any time to improve results further. 

\begin{table}[t]
\begin{center}
\input{Tables/FusionNet_results}
\end{center}
\vspace*{3mm}
\caption{Comparison of FusionNet to the state-of-the-art. The upper section of the table corresponds to the input methods used for FusionNet. FusionNet performs better than any of the input methods. The Oracle Fusion refers to a fusion based on the ground-truth error. 
\label{tab:fusiontable}
}
\end{table}
\setlength{\tabcolsep}{1.4pt}

Table \ref{tab:fusiontable} compares FusionNet to the state-of-the-art on the common benchmark datasets.
FusionNet consistently outperforms each of the techniques that have been provided as input, which demonstrates that the assessment network is able to locally select the best optical flow vectors.
As a consequence, this brings it close to the most recent state of the art and would most likely outperform it if these methods were also included for selection.
Table~\ref{tab:fusiontable} also reports the results when selecting the flow vectors based on the ground-truth error. This oracle fusion is the lower bound that FusionNet can achieve with the respective optical flow fields given as input. 

\begin{figure}
  \resizebox{\linewidth}{!}{%
  \setlength{\tabcolsep}{0.7pt}%
  \begin{tabular}{cccc}
      \includegraphics[width=0.3\textwidth]{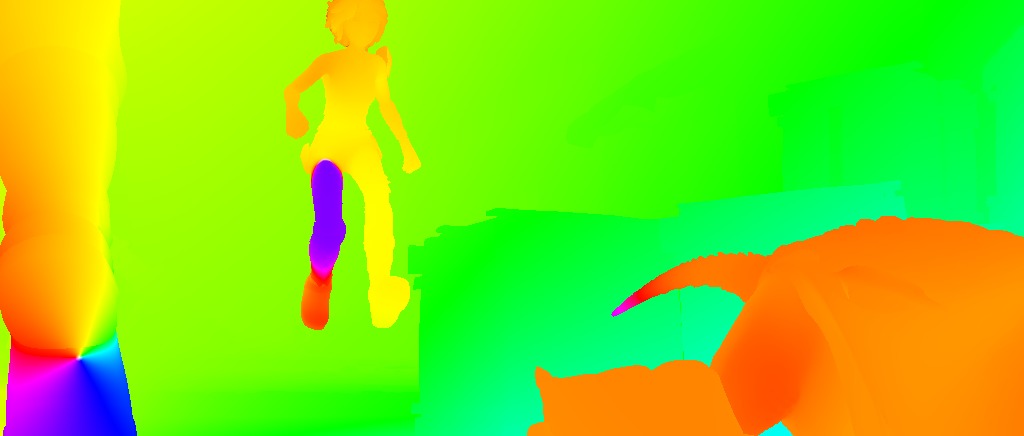}  & 
      \includegraphics[width=0.3\textwidth]{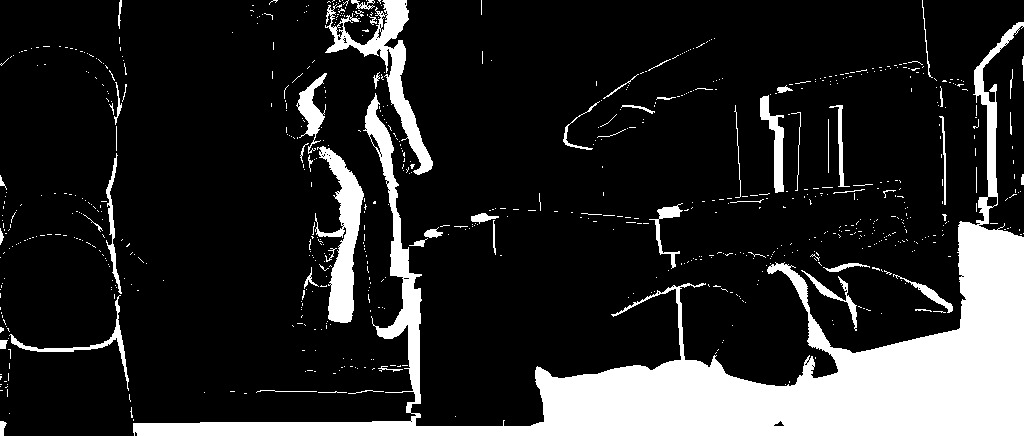} &
      \includegraphics[width=0.3\textwidth]{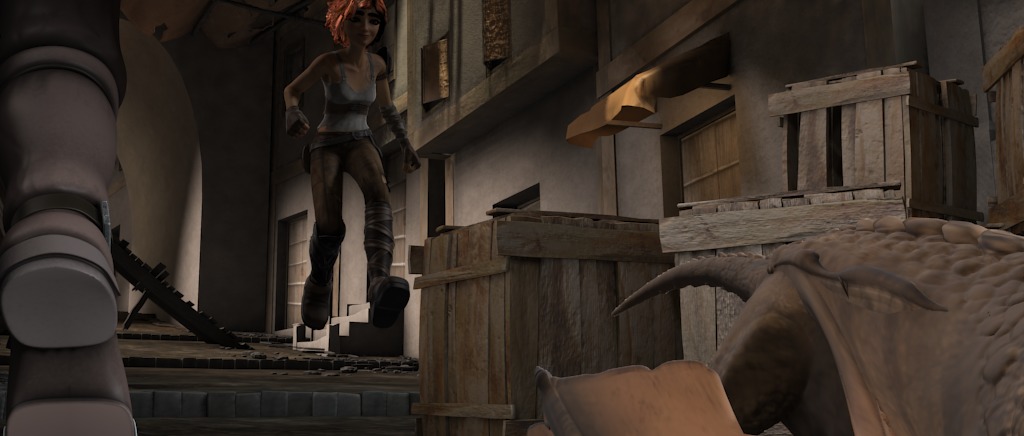} & 
      \includegraphics[width=0.3\textwidth]{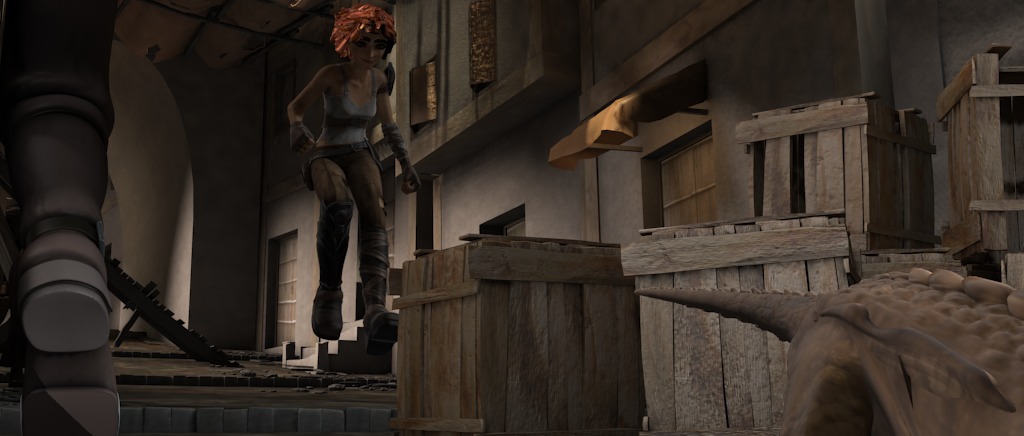}  
      \\
    Flow GT & Occlusion GT & Image 0 & Image 1 \\
    \hline \\ [-1.5ex]
      \includegraphics[width=0.3\textwidth]{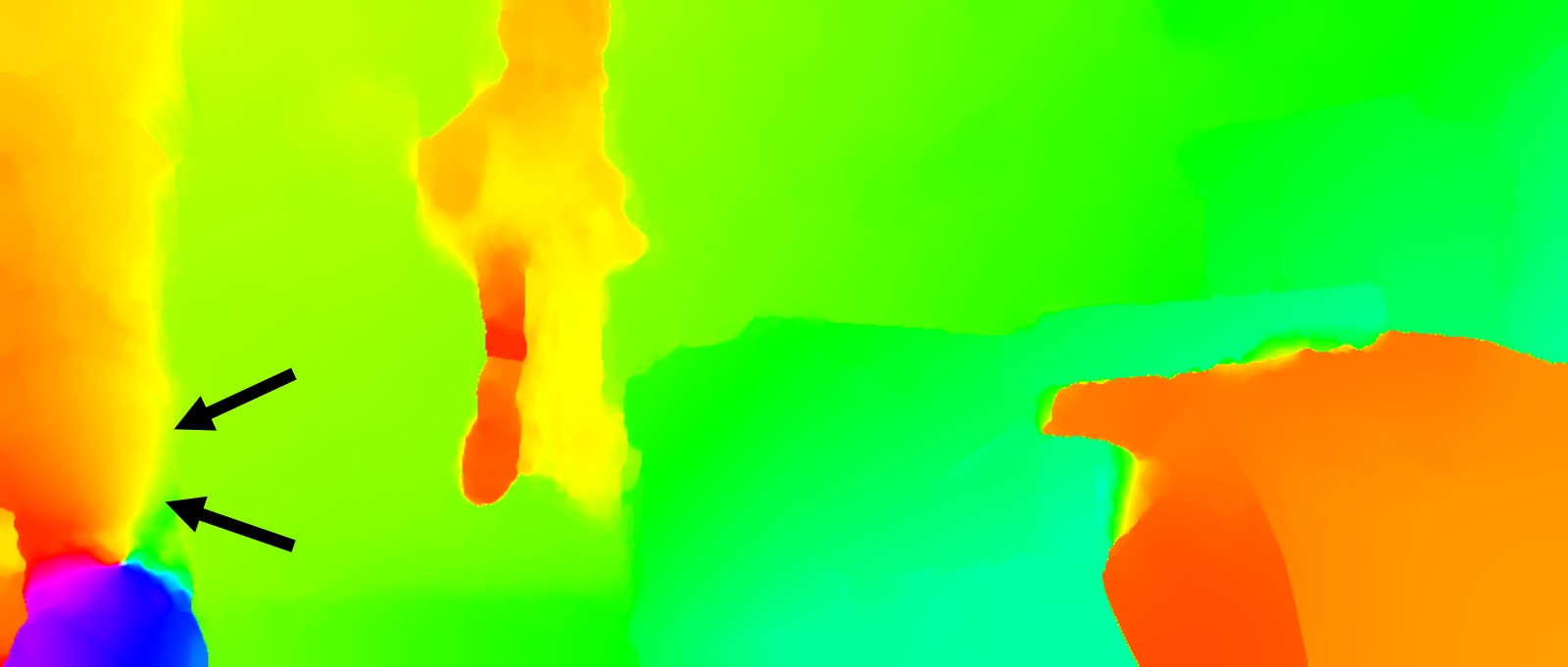} & 
      \includegraphics[width=0.3\textwidth]{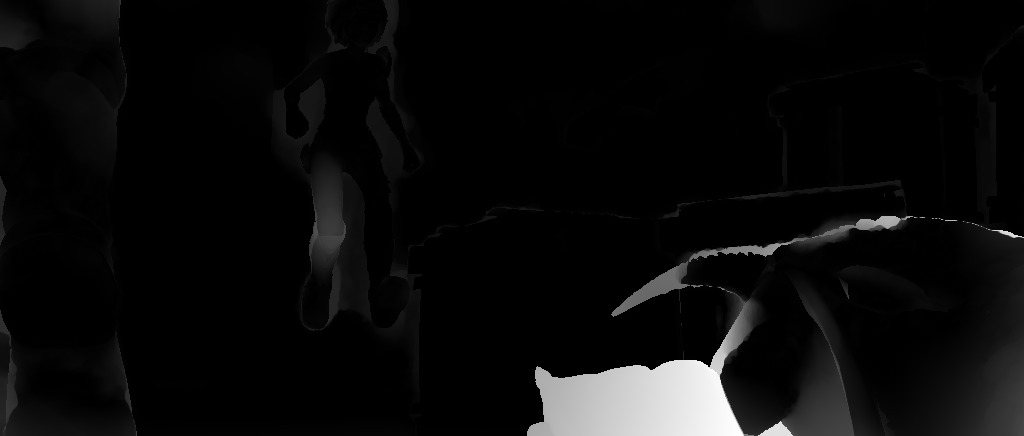} & 
      \includegraphics[width=0.3\textwidth]{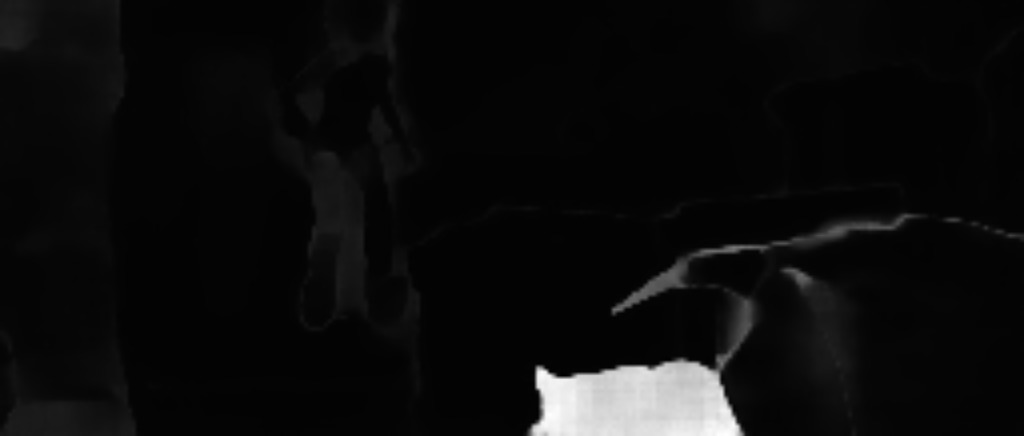} & 
      \includegraphics[width=0.3\textwidth]{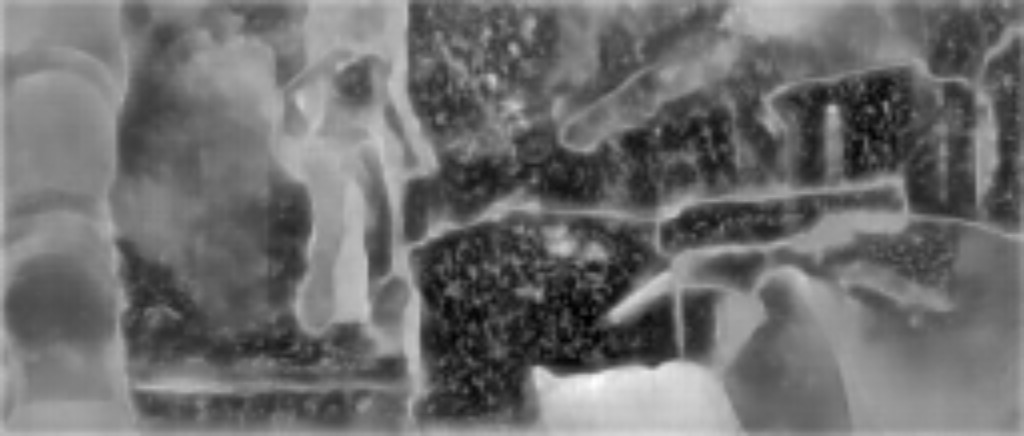} \\
    FlowFields~\cite{flowfields} flow & GT error & Predicted error (L1)  & Predicted error (Hinge) \\
      \includegraphics[width=0.3\textwidth]{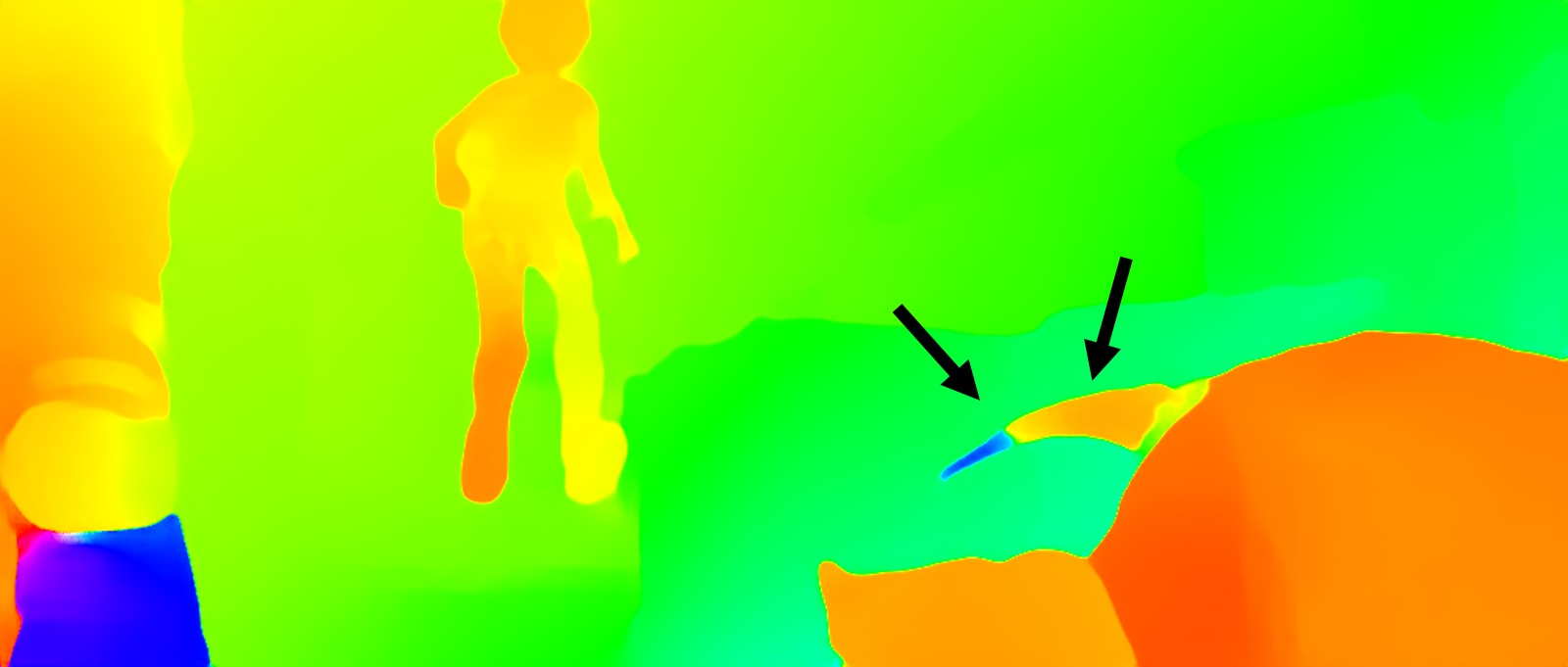} & 
      \includegraphics[width=0.3\textwidth]{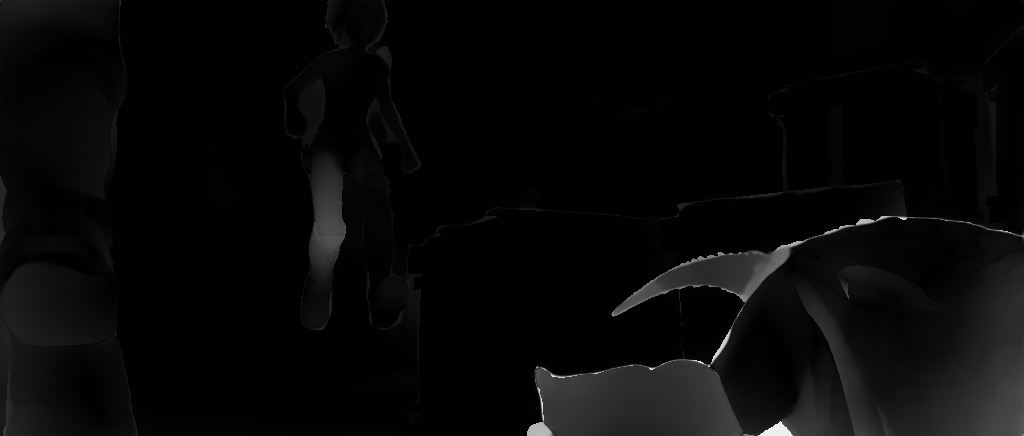} & 
      \includegraphics[width=0.3\textwidth]{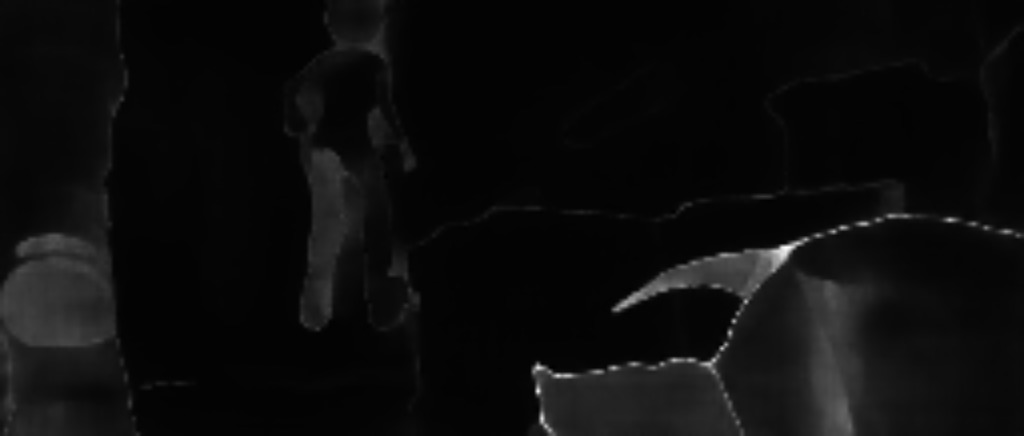} & 
      \includegraphics[width=0.3\textwidth]{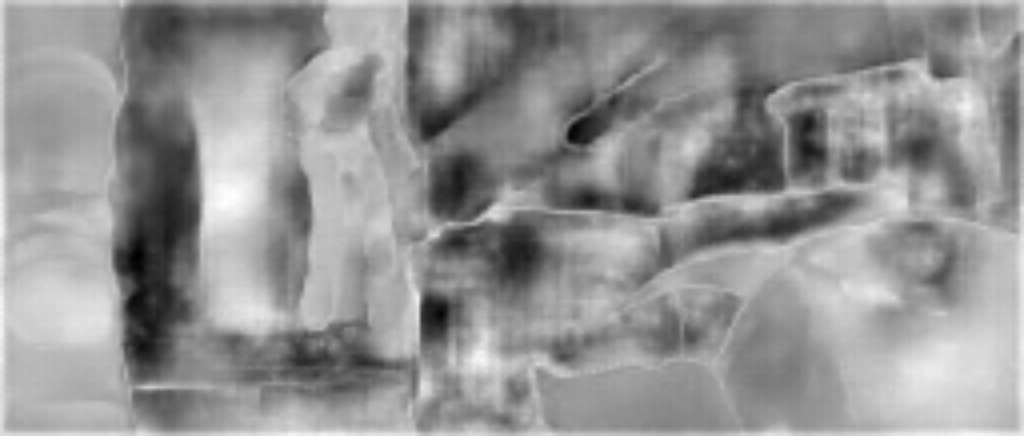} \\
    FlowNet2~\cite{flownet2} flow & GT error & Predicted error (L1) & Predicted error (Hinge) \\
      \includegraphics[width=0.3\textwidth]{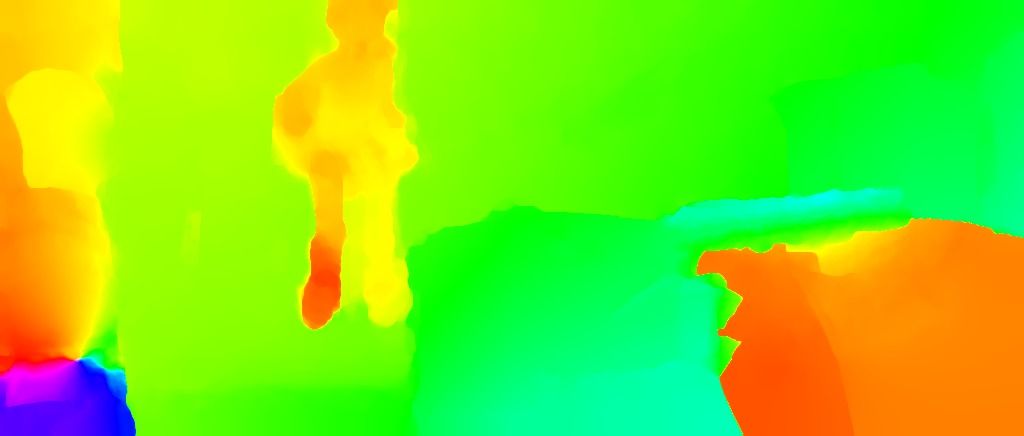} & 
      \includegraphics[width=0.3\textwidth]{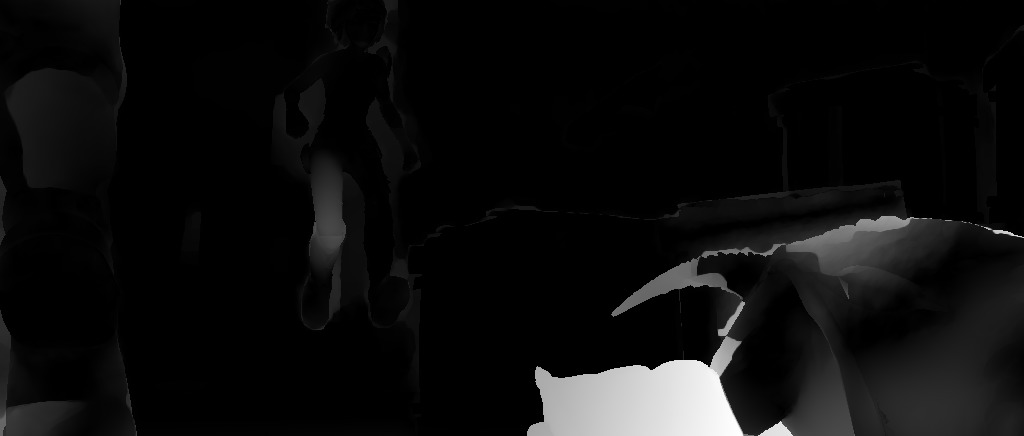} & 
      \includegraphics[width=0.3\textwidth]{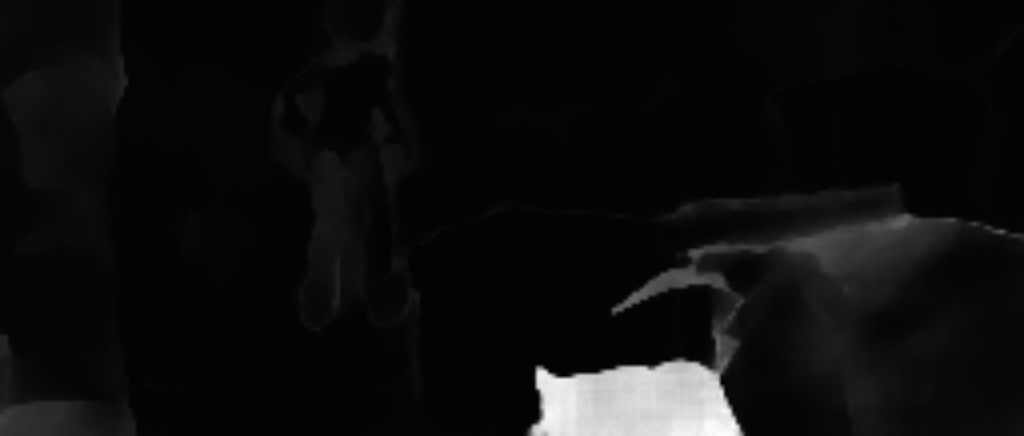} & 
      \includegraphics[width=0.3\textwidth]{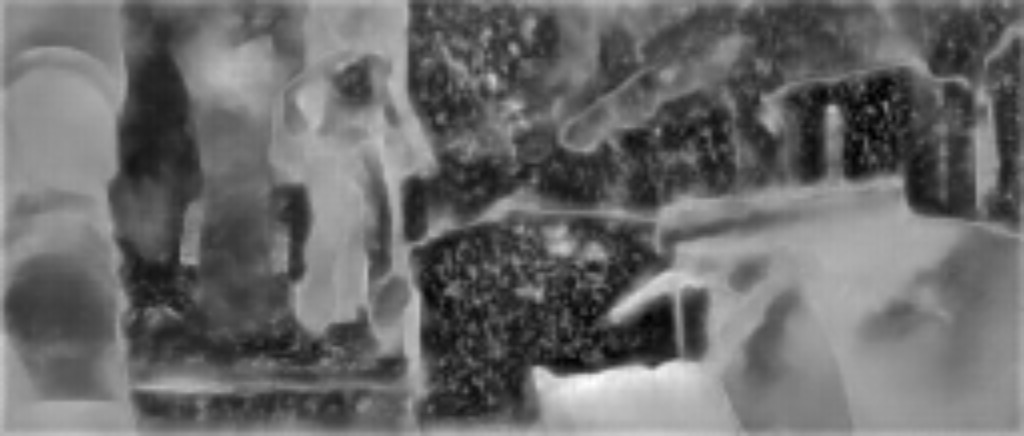} \\
    EpicFlow~\cite{epicflow} flow  & GT error & Predicted error (L1) & Predicted error (Hinge) \\
      \includegraphics[width=0.3\textwidth]{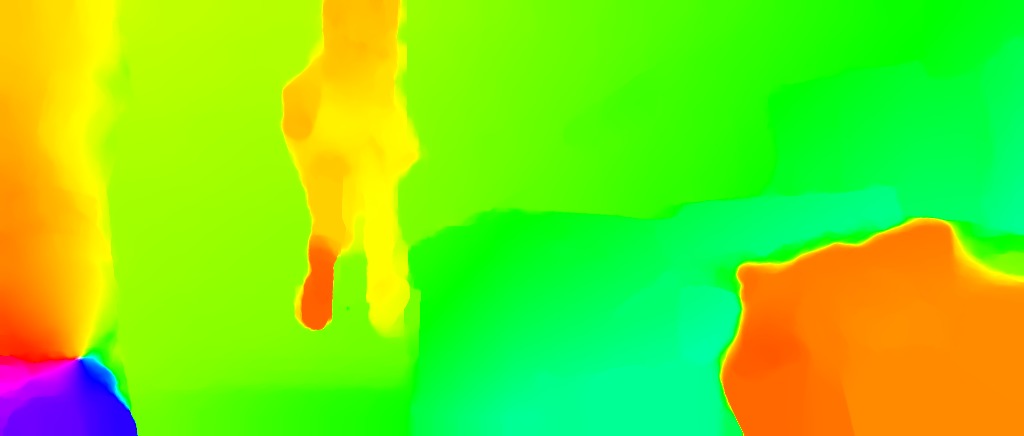} & 
      \includegraphics[width=0.3\textwidth]{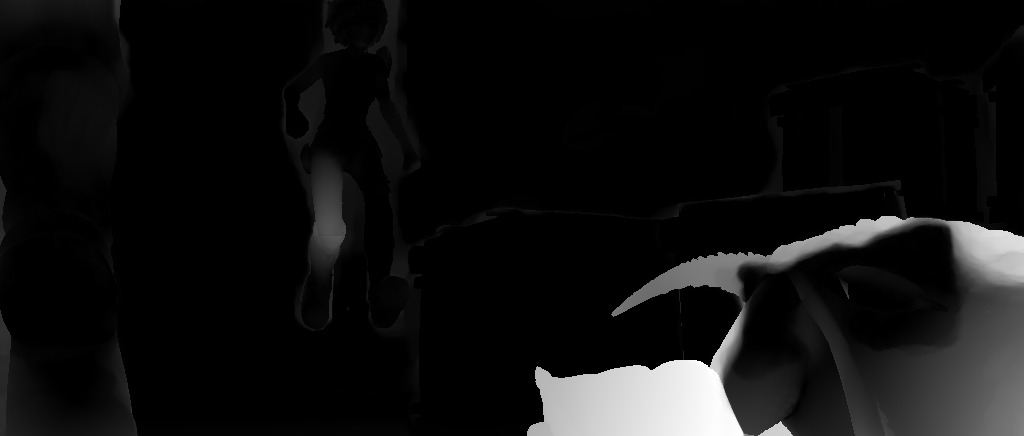} & 
      \includegraphics[width=0.3\textwidth]{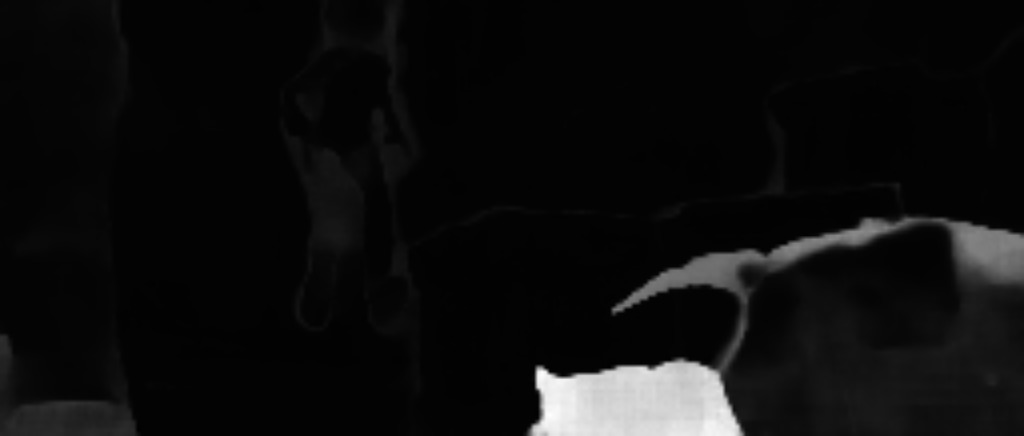} & 
      \includegraphics[width=0.3\textwidth]{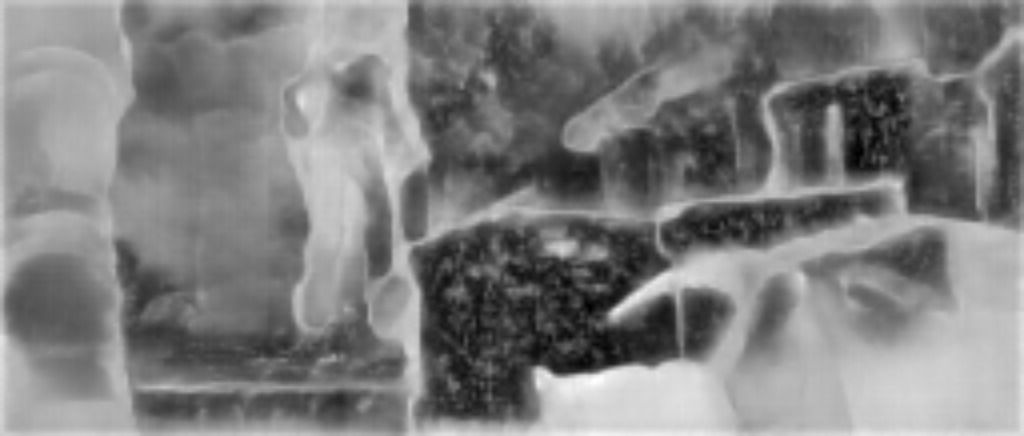} \\
    DeepFlow~\cite{deepmatching} flow &  GT error & Predicted error (L1) & Predicted error (Hinge) \\
      \includegraphics[width=0.3\textwidth]{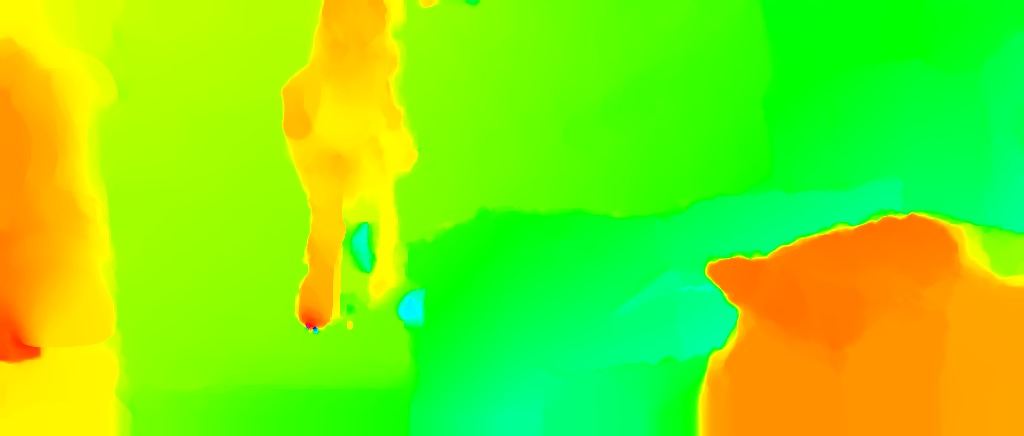} & 
      \includegraphics[width=0.3\textwidth]{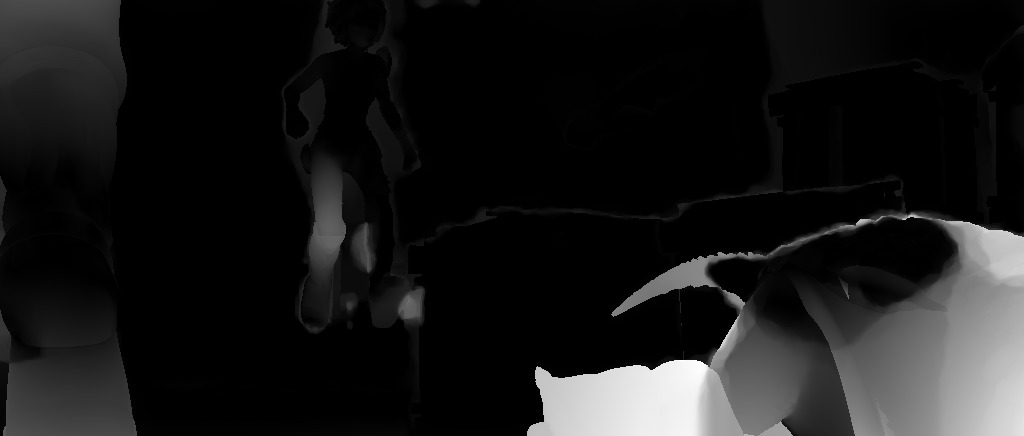} & 
      \includegraphics[width=0.3\textwidth]{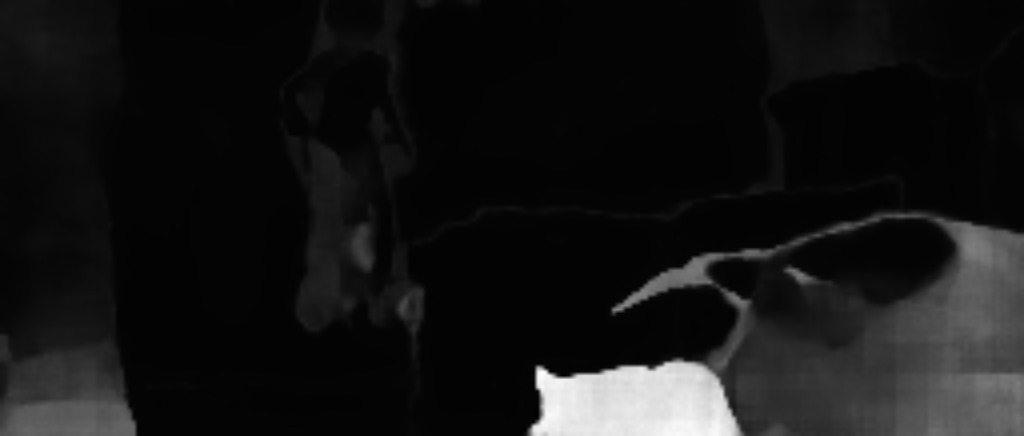} & 
      \includegraphics[width=0.3\textwidth]{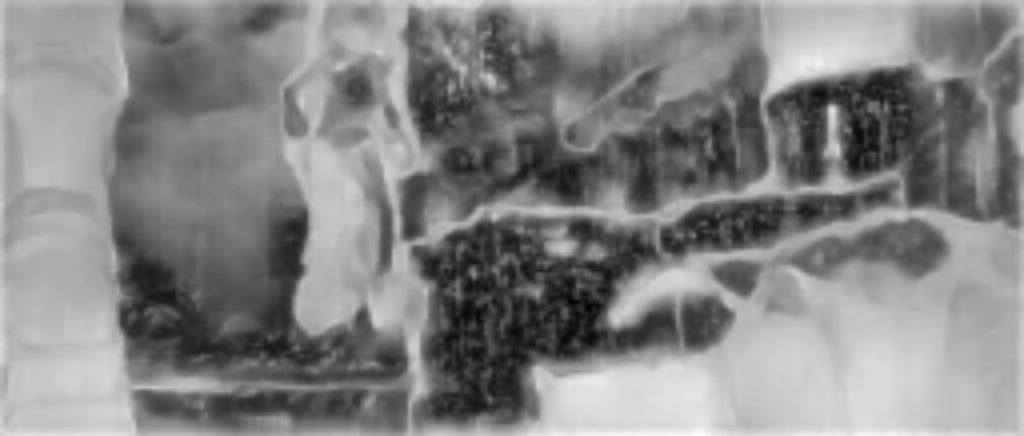} \\
    LDOF~\cite{ldof} flow & GT error & Predicted error (L1) & Predicted error (Hinge) \\
        \hline \\ [-1.5ex]
      & 
      \includegraphics[width=0.3\textwidth]{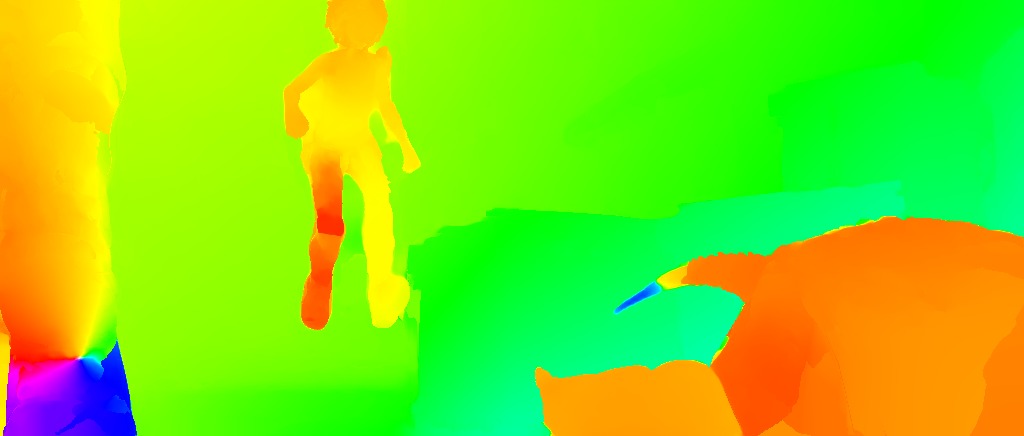} & 
      \includegraphics[width=0.3\textwidth]{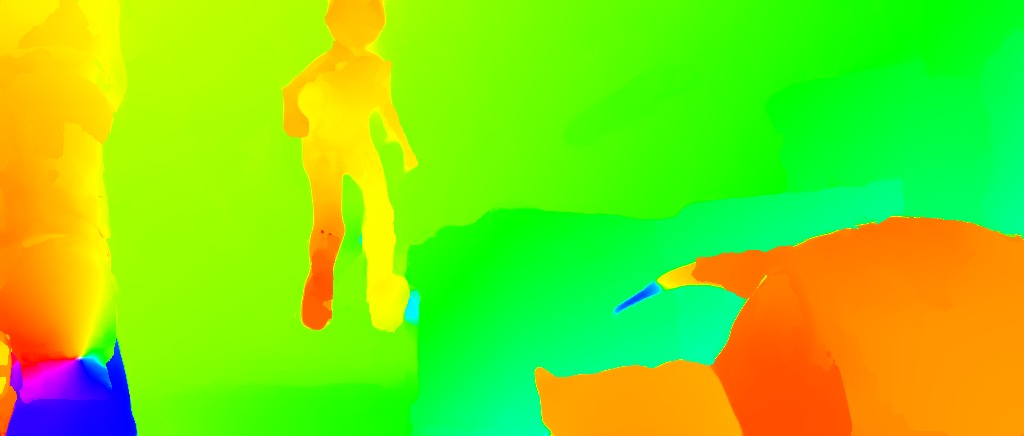} & 
      \includegraphics[width=0.3\textwidth]{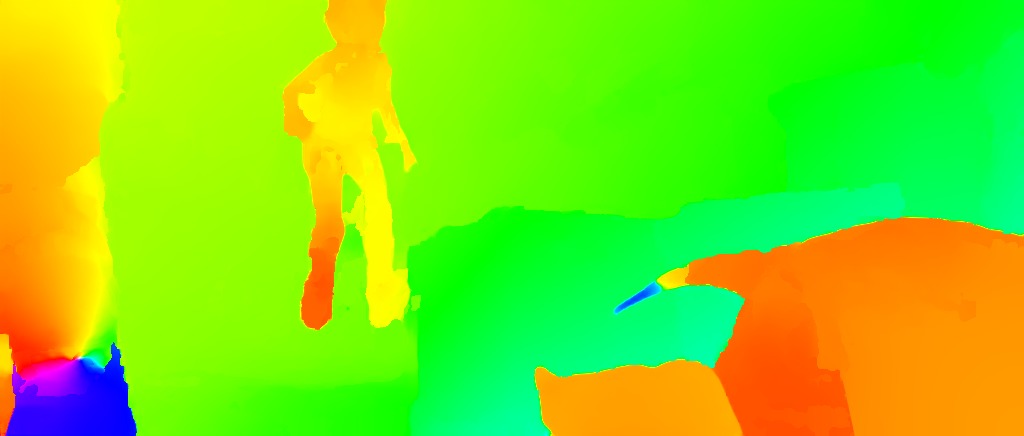} \\
    & Oracle & FusionNet (L1) & FusionNet (Hinge) \\    
\end{tabular}%
}%
     \caption{        
        Qualitative example showing the error prediction of the assessment network and the corresponding fused optical flow field. 
        \textbf{First column:} Different flows provided as input. \textbf{Second column:} Their ground truth error. \textbf{Third and fourth column:} Error prediction with the L1 loss and the hinge loss, respectively. The error predicted with the L1 loss is very close to the ground-truth error. The error predicted with the hinge loss is not interpretable, since it is a relative error and therefore is locally scaled. The bottom row shows the merged flow using the ground truth (Oracle) or the error predictions. The marked regions indicate how FusionNet picks the best motion vectors from the different input methods. 
        \label{fig:qualitative_single_ass}
        } 
\end{figure}

The predicted error when trained with an L1 loss is shown in 
Figure~\ref{fig:qualitative_single_ass}. One can observe that it matches the ground-truth error quite well. Thus, in hard cases, if one of the input methods is able to estimate the motion successfully, FusionNet is able to select the best estimate. The predicted error for the margin loss is not directly interpretable due to the local scaling. However, the resulting final optical flow field is as good as the one obtained with the L1 loss. Also quantitatively comparing the L1 loss against the hinge loss, there is no significant difference.

\subsection{Augmented FlowNet}

While FusionNet yields excellent optical flow that combines the best from all available methods, it requires 84 seconds per frame. In contrast, FlowNet 2.0 runs at 8 frames per second \footnote{FlowNet2 runtime is reported on an Nvidia GTX1080 GPU, while the classical methods run on the CPU.}. In this section we test how far we can transfer the good results from FusionNet to a FlowNet, thus inheriting also the runtimes of the latter. 

Table~\ref{tab:AugmentedFlowNetC} first shows the influence of the choice of the proxy ground-truth when fine-tuning a basic FlowNetC. 
Augmenting the FlowNet with an optical flow field that is superior to the baseline improves results, whereas inferior flow fields can decrease the performance. When using just a single proxy method, there is the dilemma of which method to choose. 
Feeding a random mixture of samples from various methods during fine-tuning (Rand. Mix) does not yield the best of all involved methods, but approximately the average of those. In contrast, the use of FusionNet resolves the dilemma.

\setlength{\tabcolsep}{4pt}
\begin{table}[t]
\begin{center}
\input{Tables/AugmentedFlowNetC_results}
\end{center}
\vspace*{3mm} 
\caption{Influence of the proxy ground truth on the augmented FlowNetC. Average endpoint errors on the training set of Sintel and KITTI are reported. Augmentation with a single proxy can improve results, but it is not obvious which method to choose. Using multiple methods to generate the proxy ground-truth (FusionNet) yields consistent improvements across all benchmarks. The benefits come largely from combining FlowNet2 and FlowFields. The upper part of the table shows experiments which are trained on domain-specific data (denoted AugmentedFlowNetD), while the bottom part shows experiments where the training data came from multiple domains to yield a generic network (denoted AugmentedFlowNetG). 
\label{tab:AugmentedFlowNetC}
}
\end{table}
\setlength{\tabcolsep}{1.4pt}

We also distinguish between augmentation for a specific domain and generic augmentation. In the first case, we augment the FlowNet only on data from the respective domain, i.e., animation movies in case of Sintel and driving videos in case of KITTI; in the second case, data from both domains is used for finetuning. Specialization to a certain domain is one of the big advantages of learning-based optical flow methods, and a particular advantage for those methods that do not require supervision in that domain, as in our case. 

Table~\ref{tab:AugmentedFlowNetC} shows that domain-specific augmentation improves results considerably on KITTI, which is a very special scenario. The error is almost cut into half. However, also the generic augmentation is not much worse, as it also benefits from the training data from the special domain, even though it is now mixed with data from another domain. Obviously, the network can figure out automatically at test time from which domain the input is from, and applies the appropriate priors from that domain.

Table~\ref{tab:AugmentedFlowNetStacks} extends the augmentation to a stacked FlowNet and compares it to UnFlow~\cite{unflow}. UnFlow uses an unsupervised loss, thus it can be specialized conveniently to any domain. The table shows results for UnFlow trained on CityScapes or the unlabeled data from KITTI, which outperform the supervised baseline, which was trained on synthetic data outside this domain. For better comparison to our strategy, we also report results for a semi-supervised version of UnFlow, i.e., it is initialized with a FlowNet trained on synthetic data before the unsupervised training starts. 

\definecolor{LightGray}{rgb}{0.9,0.9,0.9}
\setlength{\tabcolsep}{4pt}
\begin{table}[t]
\begin{center}
\input{Tables/AugmentedFlowNetStacks_results}
\end{center}
\vspace*{3mm} 
\caption{Comparison of augmented FlowNet stacks to UnFlow~\cite{unflow}. AugmentedFlowNetD-C and UnFlow-C-ours are trained on the same domain-specific data and are initialized with the same model (Baseline). The results show that the purpose of domain adaptation is better achieved with the augmentation based on FusionNet  than with the unsupervised loss of UnFlow. The results for the FlowNet augmented with data from both domains even show that it is not necessary to train separate networks for each domain, but that a generic network augmented on both domains is equally good. 
\label{tab:AugmentedFlowNetStacks}
}
\end{table}
\setlength{\tabcolsep}{1.4pt}

Results show that the domain adaptation with the augmented FlowNet is clearly superior to the one of UnFlow\footnote{UnFlow does not require any supervision, which makes it biologically more plausible. From the engineering perspective, however, this is irrelevant.}. As we already observed in Table~\ref{tab:AugmentedFlowNetC}, there is no significant difference between domain-specific training and training on a joint set of domains. This is also true for the stacked network.  

\setlength{\tabcolsep}{4pt}
\begin{table}[t]
\begin{center}
\input{Tables/Benchmark_results}
\end{center}
\vspace*{3mm}
\caption{Comparison to the state-of-the-art. Numbers marked with $^\dagger$ have been obtained after fine-tuning on the training set of the respective benchmark. On the KITTI benchmarks, we clearly extend the state of the art. Thanks to additional fine-tuning with ground truth data, the augmented network even performs better than the FusionNet proxy, but also the generic version, which has only been finetuned on the FusionNet proxy, gets close to FusionNet and sometimes even outperforms it.
\label{tab:Benchmark_results}
}
\end{table}
\setlength{\tabcolsep}{1.4pt}

Table~\ref{tab:Benchmark_results} compares the stacked augmented FlowNet to the state of the art. 
On the KITTI benchmarks, the augmented FlowNet sets the new state of the art after being finetuned also with the ground truth from the KITTI training set. But also the generic version, which has not been finetuned with ground truth data yields very good results. The direct comparison to FlowNet2 quantifies the improvement on stacked networks due to the augmentation. 

Interestingly, the stacked augmented FlowNet often even outperforms the FusionNet proxy. This is due to the finetuning with ground truth in case of the domain-specific network. Sometimes, also the generic network is better than FusionNet, but not consistently. 

\begin{figure}
 \resizebox{\linewidth}{!}{%
  \setlength{\tabcolsep}{0.7pt}%
  \begin{tabular}{ccc}%
      \vspace{-1mm}%
      \includegraphics[width=0.3\textwidth]{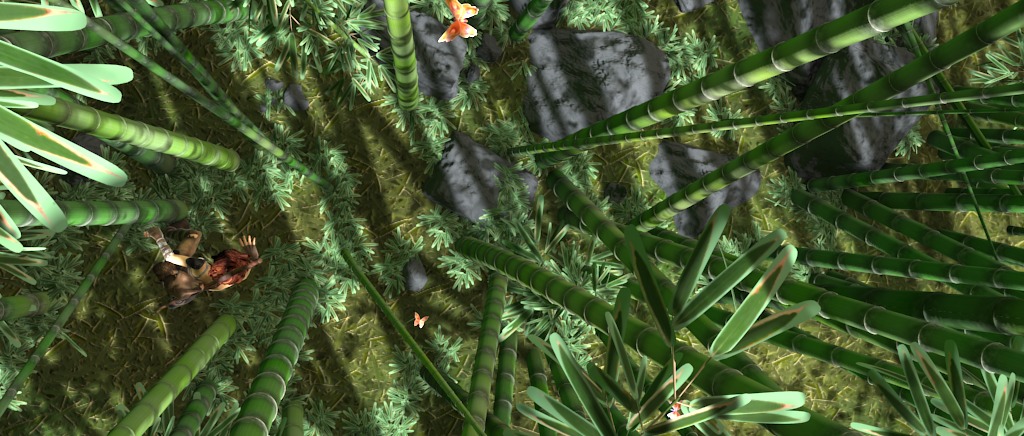} & 
      \includegraphics[width=0.3\textwidth]{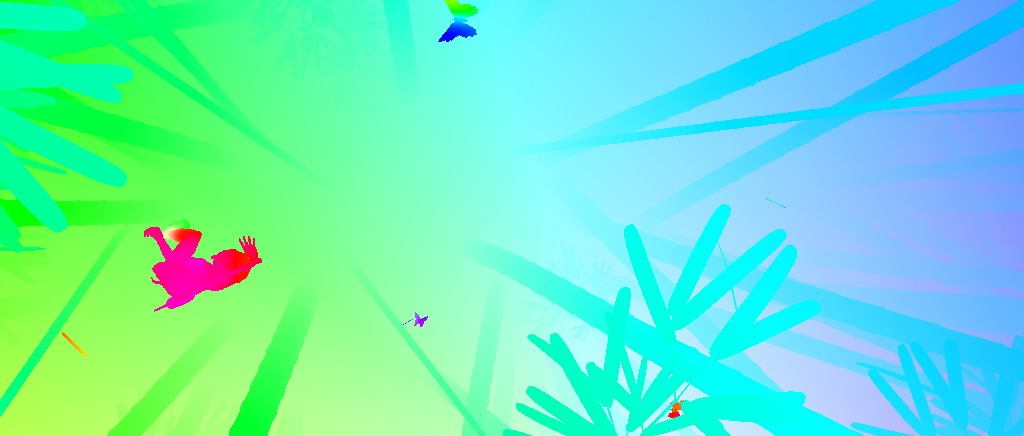} &
      \includegraphics[width=0.3\textwidth]{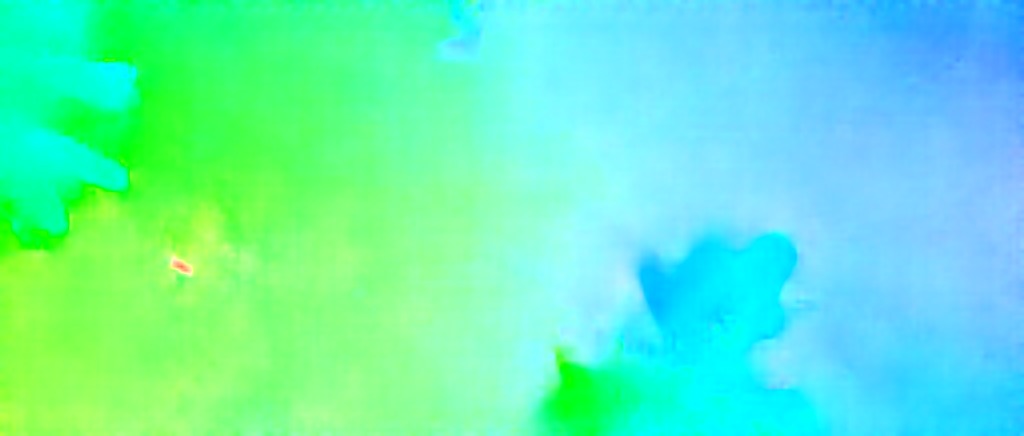} 
      \\
      {\tiny Image 1} & {\tiny Flow GT} & {\tiny Baseline Flow}\\
      \vspace{-1mm}%
      \includegraphics[width=0.3\textwidth]{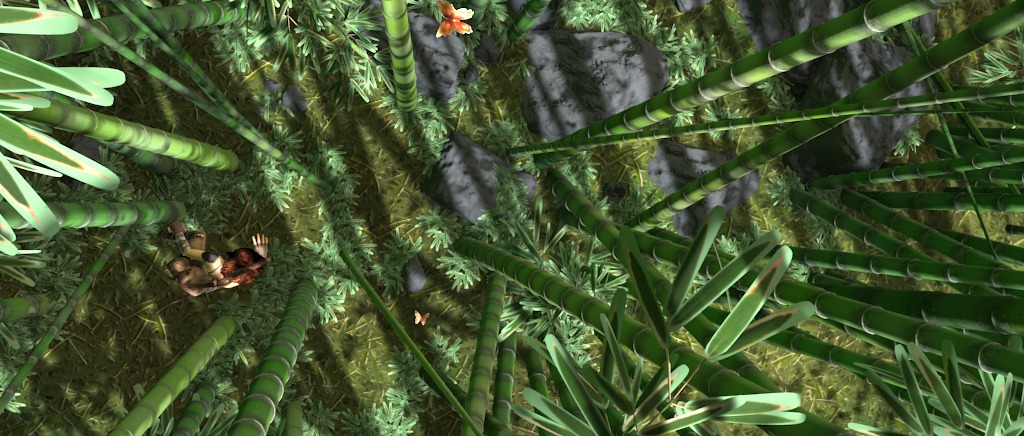} & 
      \includegraphics[width=0.3\textwidth]{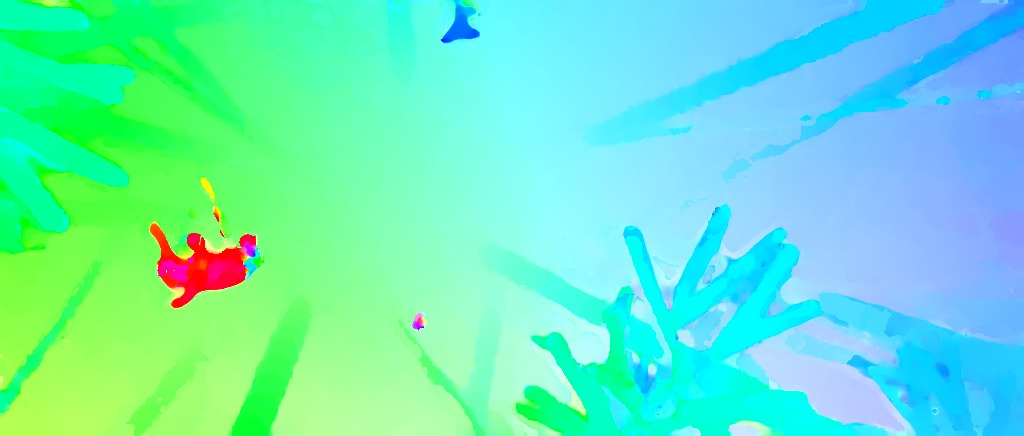} & \includegraphics[width=0.3\textwidth]{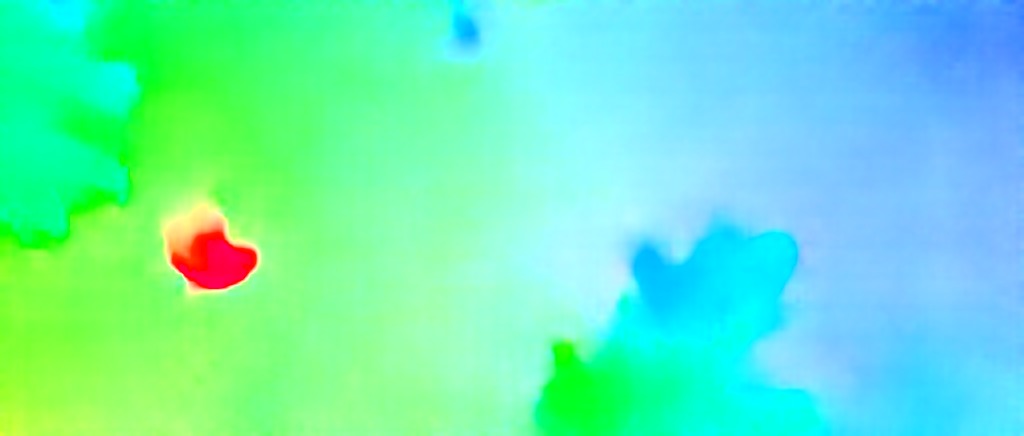}
      \\
      \vspace{5mm}%
      {\tiny Image 2} & {\tiny FusionNet Proxy} & {\tiny Augmented Flow}\\
      \vspace{-1mm}%
      \includegraphics[width=0.3\textwidth]{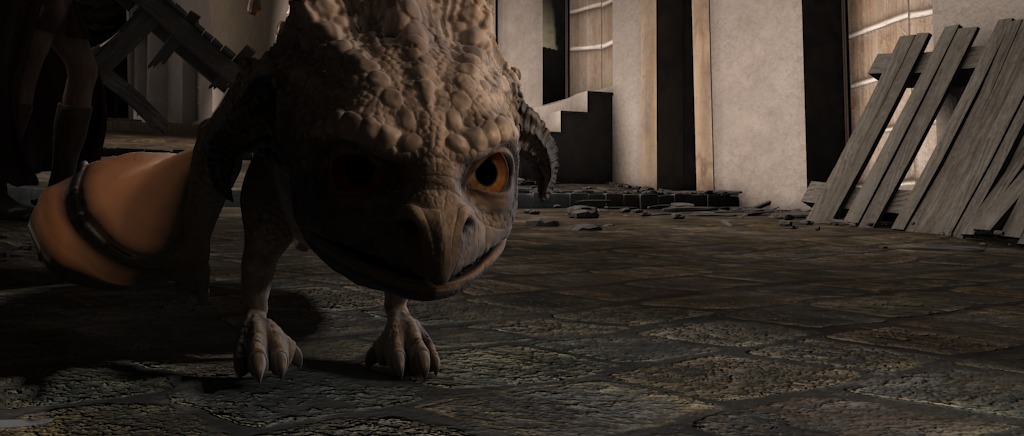} & 
      \includegraphics[width=0.3\textwidth]{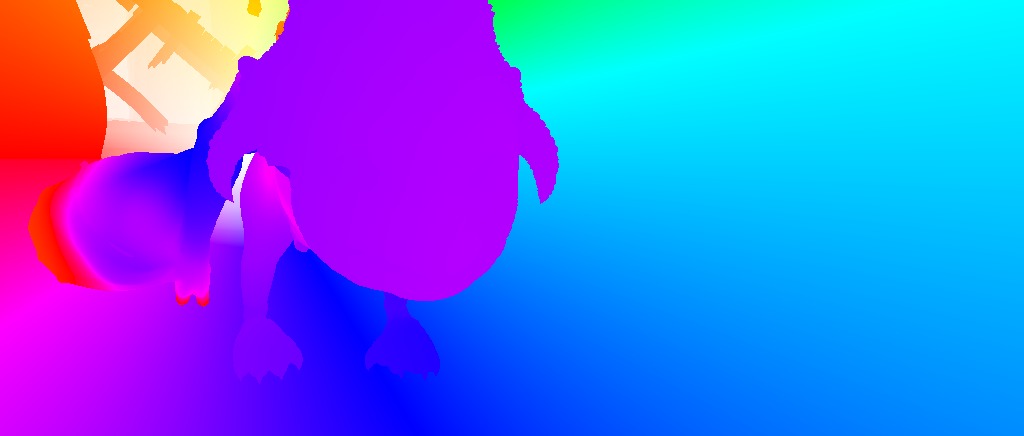} &
      \includegraphics[width=0.3\textwidth]{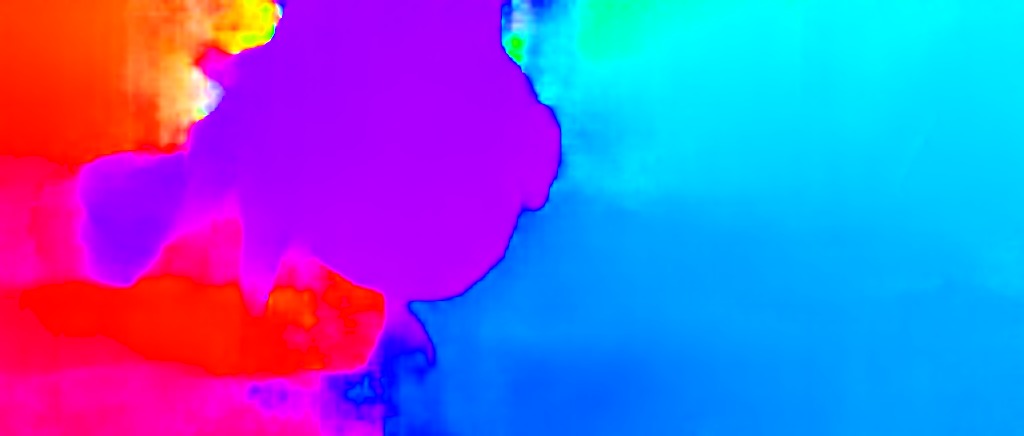} 
      \\
      {\tiny Image 1} & {\tiny Flow GT} & {\tiny Baseline Flow}\\
      \vspace{-1mm}%
      \includegraphics[width=0.3\textwidth]{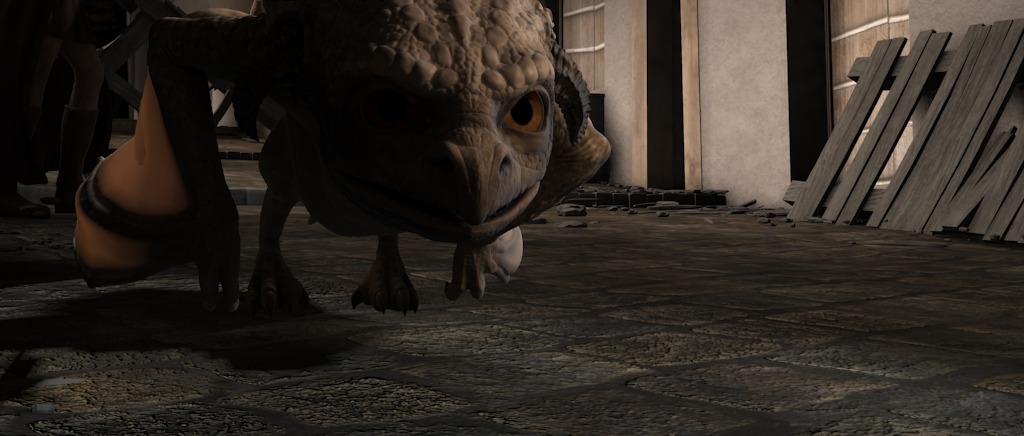} & 
      \includegraphics[width=0.3\textwidth]{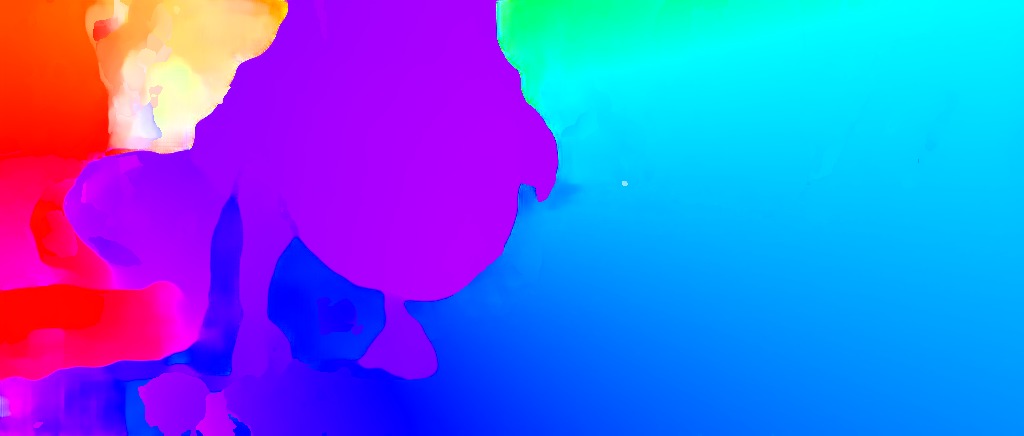} &  \includegraphics[width=0.3\textwidth]{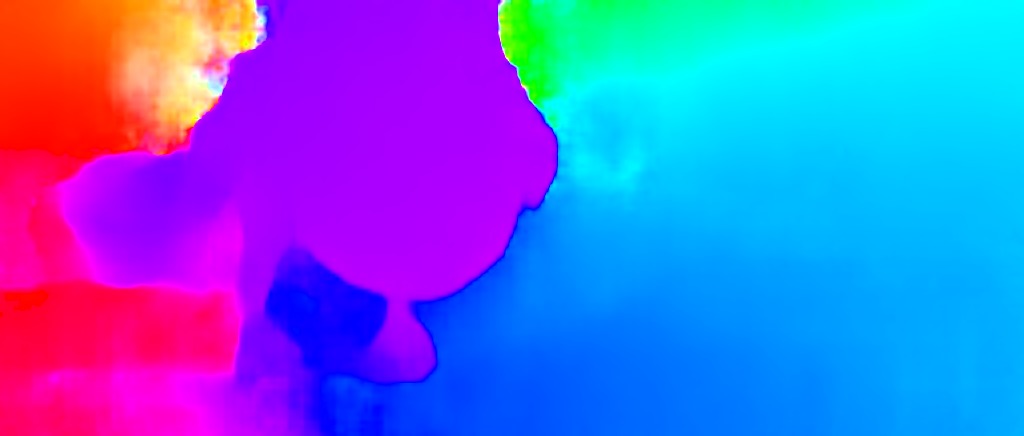}
      \\
      \vspace{5mm}%
      {\tiny Image 2} & {\tiny FusionNet Proxy} & {\tiny Augmented Flow}\\
      \vspace{-1mm}%
      \includegraphics[width=0.3\textwidth]{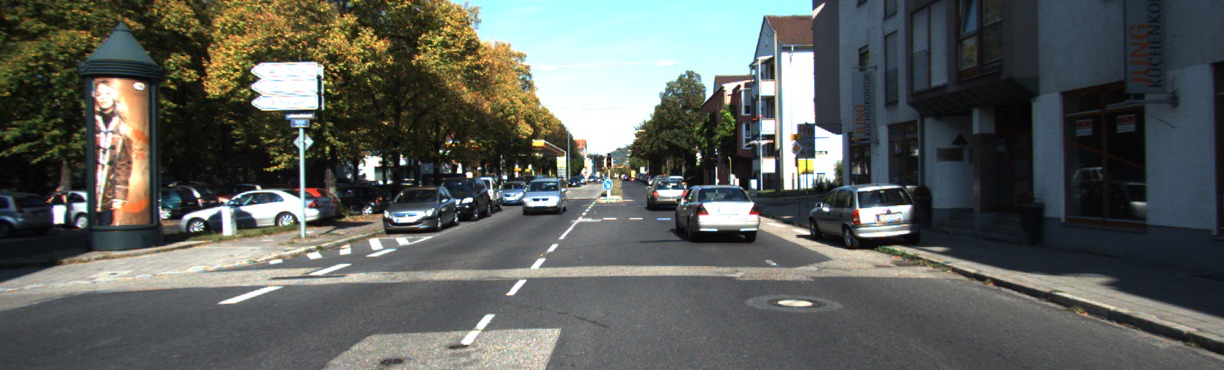} & 
      \includegraphics[width=0.3\textwidth]{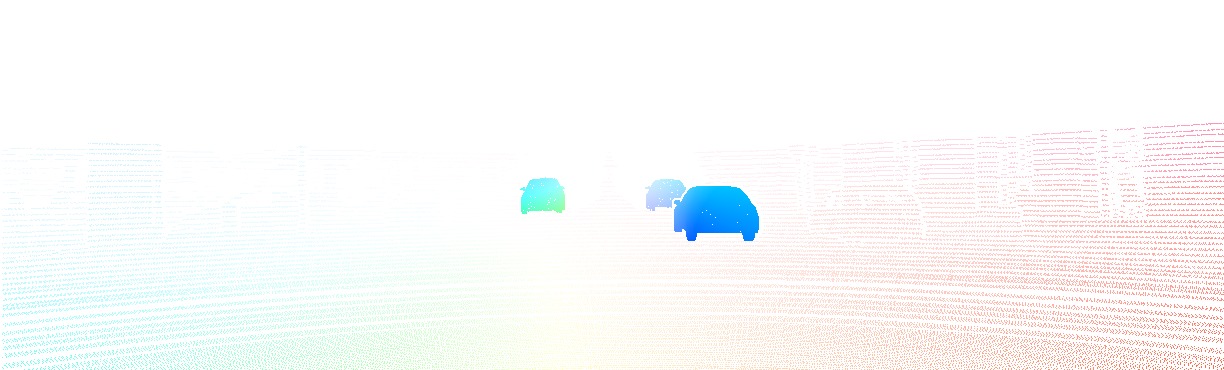} &
      \includegraphics[width=0.3\textwidth]{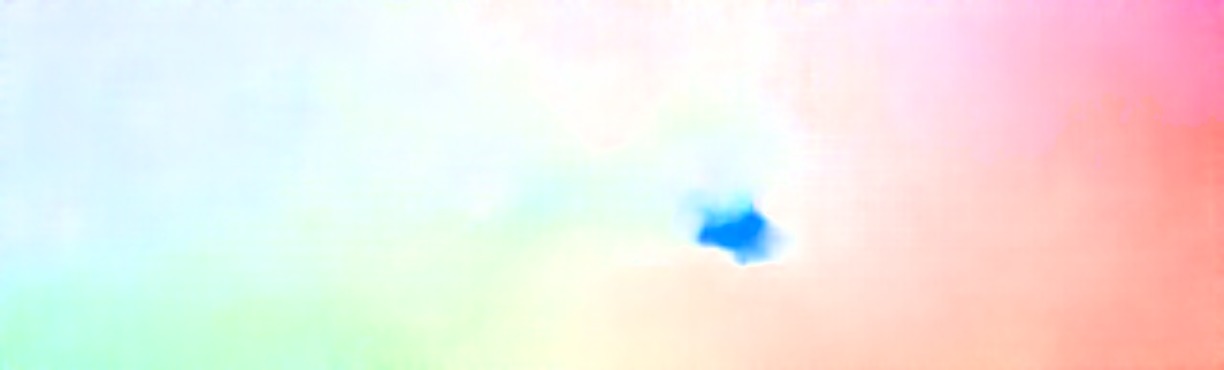}
      \\
      {\tiny Image 1} & {\tiny Flow GT} & {\tiny Baseline Flow}\\
      \vspace{-1mm}%
      \includegraphics[width=0.3\textwidth]{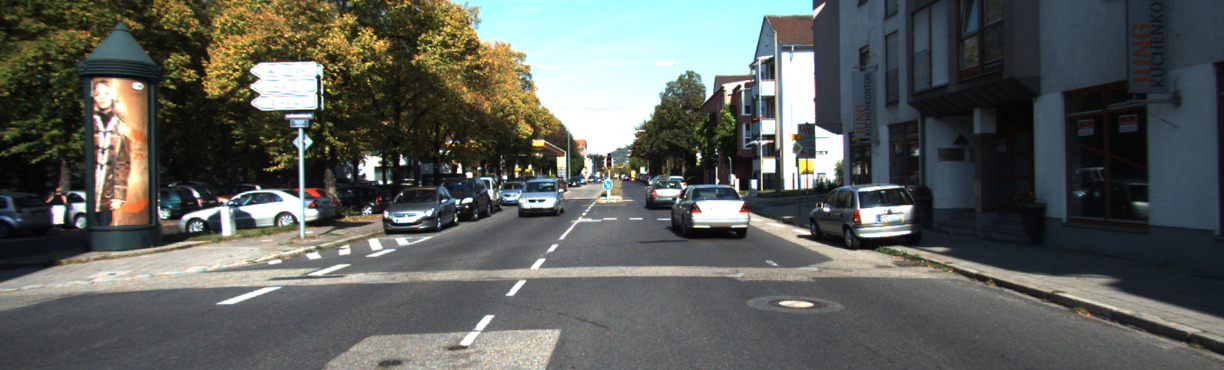} & 
      \includegraphics[width=0.3\textwidth]{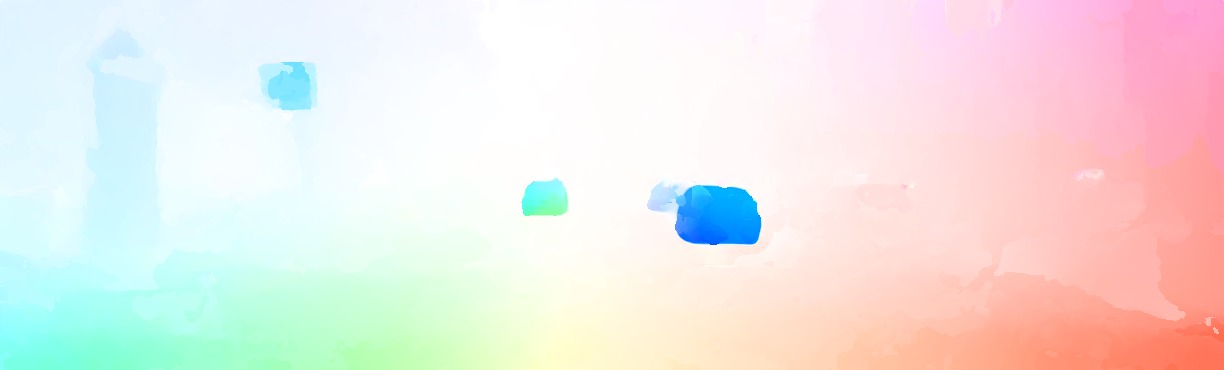}  &
      \includegraphics[width=0.3\textwidth]{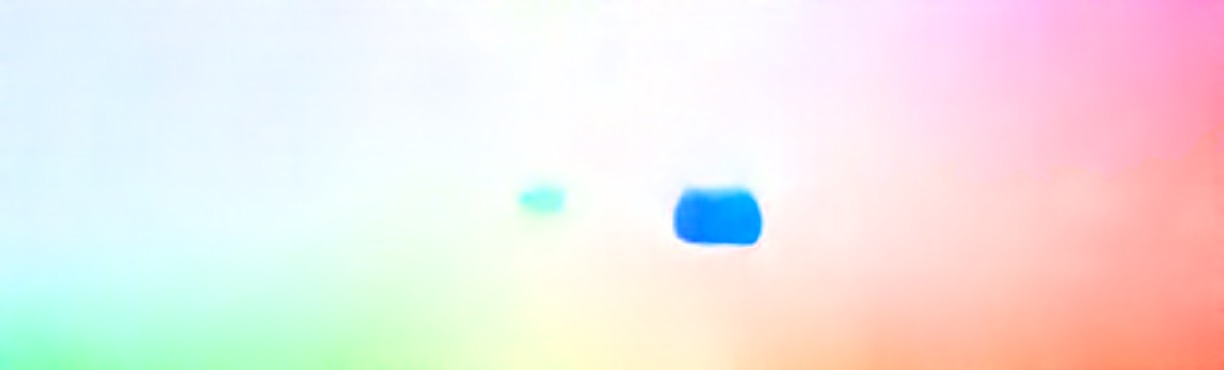} 
      \\
      \vspace{5mm}%
      {\tiny Image 2} & {\tiny FusionNet Proxy} & {\tiny Augmented Flow}\\
      \vspace{-1mm}%
      \includegraphics[width=0.3\textwidth]{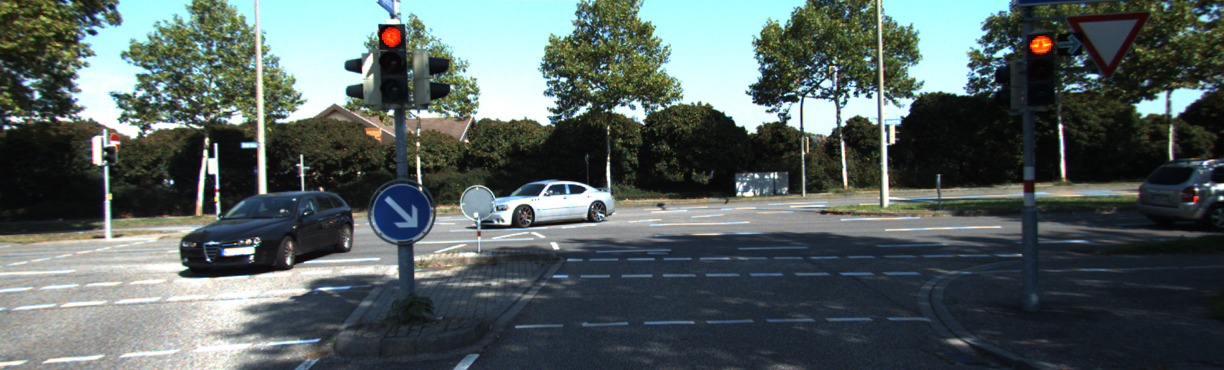} & 
      \includegraphics[width=0.3\textwidth]{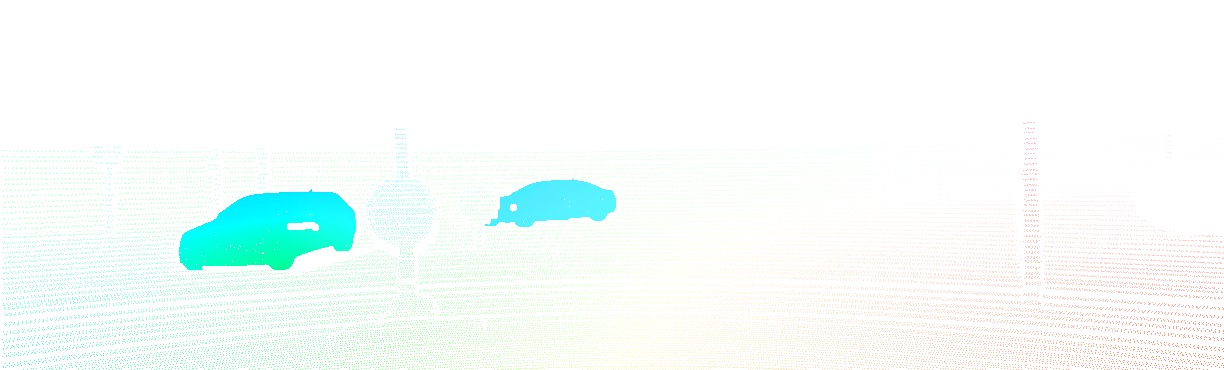} &
      \includegraphics[width=0.3\textwidth]{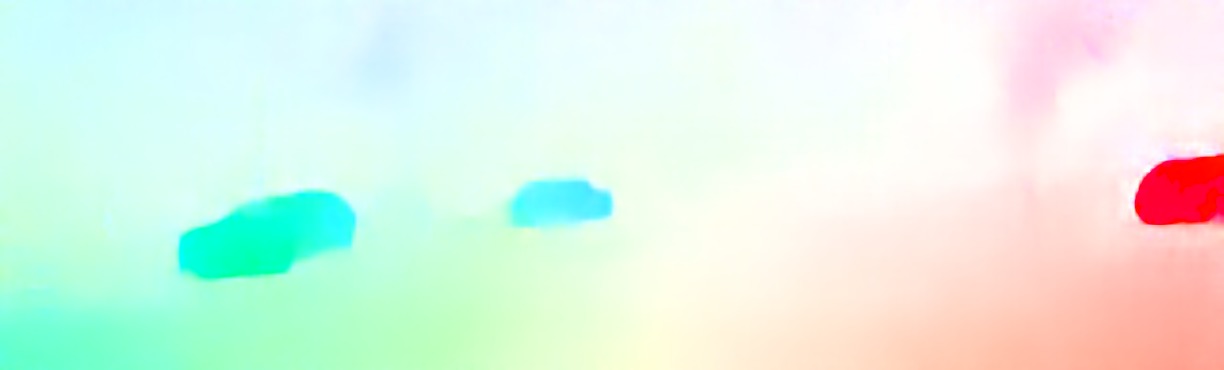} 
      \\
      {\tiny Image 1} & {\tiny Flow GT} & {\tiny Baseline Flow}\\
      \vspace{-1mm}%
      \includegraphics[width=0.3\textwidth]{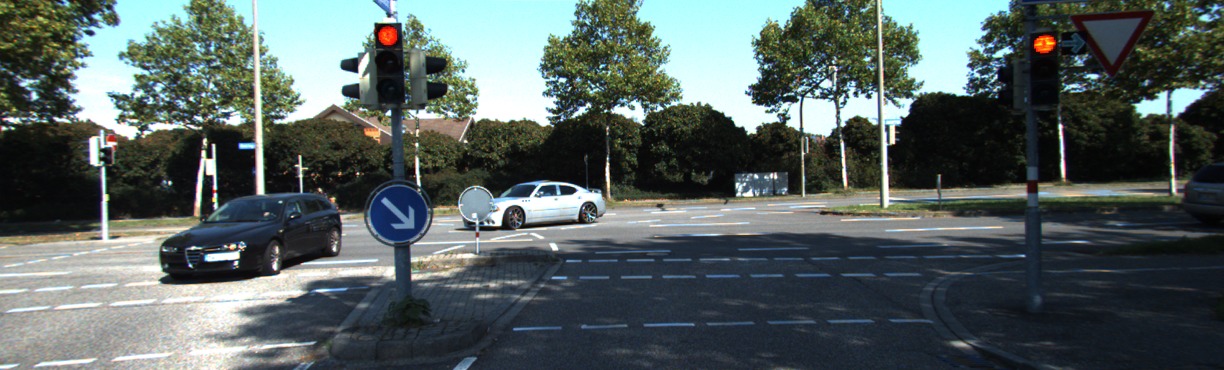} & 
      \includegraphics[width=0.3\textwidth]{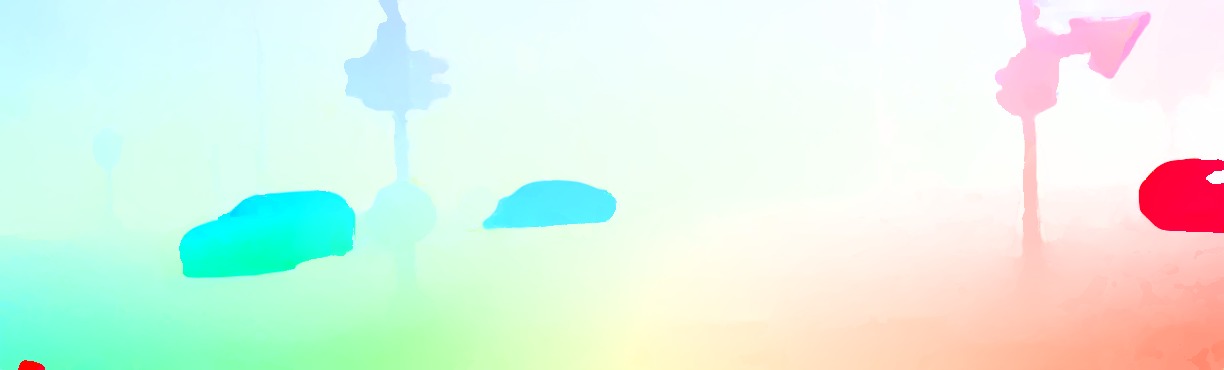} &  
      \includegraphics[width=0.3\textwidth]{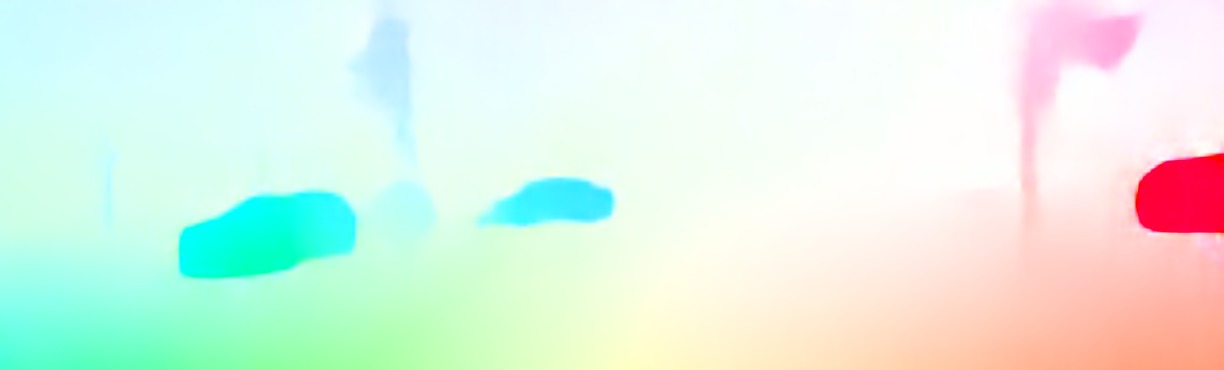} 
      \\
      {\tiny Image 2} & {\tiny FusionNet Proxy} & {\tiny Augmented Flow}\\
\end{tabular}%
}%
     \caption{Examples from Sintel Clean and KITTI2015. Using the proxy ground truth obtained with FusionNet for finetuning improves the network's performance. Note that the ground truth on KITTI is sparse and lacks some moving objects.
      \label{fig:qualitative_augmented_FlowNet_Sintel}
     } 
    \end{figure}

Fig.~\ref{fig:qualitative_augmented_FlowNet_Sintel} shows some qualitative examples of the augmentation by the proxy, yet with the smaller, non-stacked network. 

Finally, we also evaluated the augmented FlowNet in an application scenario. We augmented the FlowNet on data from UdG-MS19 and UdG-MS20~\cite{udg} and fed its optical flow into the motion segmentation approach from Keuper et al.~\cite{KB15b}. The motion segmentation performance was evaluated on the FBMS benchmark~\cite{OB14b}.
Table~\ref{tab:augmentedFlowNet_motion_seg} shows that the adaptation to real images clearly helps a small FlowNetC to improve motion segmentation results. For the larger, stacked network, we do not see a significant improvement due to the augmentation. We attribute this to some saturation effect: the optical flow of FlowNet2 is already very good, such that other parts of the motion segmentation pipeline dominate the final result. 

\setlength{\tabcolsep}{4pt}
\begin{table}[t]
\begin{center}
\input{Tables/MotionSeg_results}
\end{center}
\vspace*{3mm}
\caption{Motion segmentation results on the FBMS test set~\cite{OB14b}. We fed the optical flow from the listed methods to the motion segmentation approach from Keuper et al.~\cite{KB15b}. The augmentation on real data clearly improved the FlowNetC baseline. For the stacked network, we do not see a significant improvement. This is mainly because the optical flow is already very good in this case and other effects in the motion segmentation pipeline dominate the final result.
\label{tab:augmentedFlowNet_motion_seg}
}
\end{table}
\setlength{\tabcolsep}{1.4pt}

\section{Conclusion}

In this paper, we have presented two contributions: (1) We have presented a way to assemble a high-quality flow field from a set of input flow fields being computed in an unsupervised manner using existing optical flow estimation methods. This has been achieved by training an assessment network that learns to predict the errors of the input techniques. (2) We have shown that finetuning a FlowNet on such high-quality flow fields allows for unsupervised adaptation of the network to a specific domain. With this strategy, we obtained state-of-the-art results on the KITTI benchmarks. Moreover, we have shown that this strategy is more successful on domain adaptation than a fully unsupervised approach that does not make use of any synthetic data. 

\section*{Acknowledgements} 

We acknowledge funding by the German Research Foundation (grant \mbox{BR 3815/10-1}) and the EU Horizon 2020 project Trimbot2020. 


\bibliographystyle{splncs}
\bibliography{egbib}

\end{document}

%% file: Tables/FusionNet_results.tex
      \begin{tabular}{|c|l||cc|cc||ccc|cc|}%
              \hhline{|-|-||--|--T---T--|}%
               & 
               & \multicolumn{4}{c||}{Animation Domain}%
               & \multicolumn{5}{c|}{Driving Domain}%
               \\%
               
               & Method
               & \multicolumn{2}{c|}{Sintel \textit{clean}}%
               & \multicolumn{2}{c||}{Sintel \textit{final}}%
               & \multicolumn{3}{c|}{KITTI 2012}%
               & \multicolumn{2}{c|}{KITTI 2015}%
               \\%
               
               &
               & \multicolumn{2}{c|}{AEE}%
               & \multicolumn{2}{c||}{AEE}%
               & \multicolumn{2}{c}{AEE}%
               & F1-noc
               & AEE
               & Fl-all
               \\%
               
               &
               & \textit{train} & \textit{test}%
               & \textit{train} & \textit{test}%
               & \textit{train} & \textit{test} & \textit{test}%
               & \textit{train} & \textit{test}%
               \\%
               
            \hhline{|=|=||==|==#===|==|}%

        \multirow{5}{*}{\rotatebox[origin=c]{90}{Inputs}} &
        LDOF\cite{ldof}%
        & $4.65$  & $7.56$
        & $6.16$  & $9.12$
        & $10.26$ & $12.4$ & $21.93\%$
        & $17.71$ & $-$
        \\%

        &
        DeepFlow\cite{deepmatching}%
        & $2.66$ & $5.38$
        & $4.02$ & $7.21$
        & $\pz{5.78}$ & $\pz{5.8}$ & $\pz{7.22}\%$
        & $13.14$ & $28.48\%$
        \\%
        
        &
        EpicFlow\cite{epicflow}%
        & $2.27$ & $4.11$
        & $3.57$ & $6.28$
        & $\pz{3.52}$ & $\pz{3.8}$ & $\pz{7.88}\%$
        & $\pz\textbf{9.23}$ & $\textbf{26.29}\%$
        \\%
        
        &
        FlowNet2\cite{flownet2}%
        & $2.02$ & $3.96$
        & $3.62$ & $6.02$
        & $\pz{4.09}$ & $-$ & $-$
        & $10.06$ & $-$
        \\%
        
        &
        FlowFields\cite{flowfields}%
        & $\textbf{1.86}$ & $\textbf{3.75}$
        & $\textbf{3.40}$ & $\textbf{5.81}$
        & $\pz\textbf{3.08}$ & $\pz\textbf{3.5}$ & $\pz\textbf{5.77}\%$
        & $\pz{9.26}$ & $-$
        \\%
        
        \hhline{|-|-||--|--|---|--|}%
        
        \multirow{3}{*}{\rotatebox[origin=c]{90}{Best}} &
        DCFlow\cite{dcflow}%
        & $-$     & $3.54$
        & $-$     & $5.12$
        & $-$     & $-$    & $-$
        & $-$     & $14.86\%$
        \\%

        &
        PWC-Net\cite{PWCnet}%
        & $2.55$ & $4.39$
        & $3.93$ & $\textbf{5.04}$
        & $\pz\textbf{4.14}$ & $\pz\textbf{1.7}$ & $\pz\textbf{4.22}\%$
        & $10.35$ & $\pz\textbf{9.60}\%$
        \\%
        
        &
        MR-Flow\cite{Wulff:CVPR:2017}%
        & $\textbf{1.83}$ & $\textbf{2.53}$
        & $3.59$ & $5.38$
        & $-$    & $-$    & $-$
        & $-$    & $12.19\%$
        \\%
        
        \hhline{|-|-||--|--|---|--|}%
        
        \multirow{2}{*}{\rotatebox[origin=c]{90}{Our}} &
        FusionNet-L1%
        & $1.59$     & $-$
        & $\textbf{3.10}$ & $-$
        & $\pz{3.11}$     & $-$    & $-$
        & $\pz\textbf{8.13}$     & $-$
        \\%

        &
        FusionNet-Hinge 
        \cellcolor{gray!15}%
        & \cellcolor{gray!15}$\textbf{1.58}$ & \cellcolor{gray!15}$\mathbf{3.20}$
        & \cellcolor{gray!15}$3.18$ & \cellcolor{gray!15}$\mathbf{5.50}$
        & \cellcolor{gray!15}$\pz\textbf{2.97}$ & \cellcolor{gray!15}$\pz\mathbf{3.6}$ & \cellcolor{gray!15}$\pz\mathbf{5.50}\%$
        & \cellcolor{gray!15}$\pz{8.18}$ & \cellcolor{gray!15}$\pz\mathbf{21.44}\%$
        \\%
        
        \hhline{|-|-||--|--|---|--|}%
        
        &
        Oracle%
        & $0.96$     & $-$
        & $1.83$     & $-$
        & $\pz{2.04}$     & $-$    & $-$
        & $\pz{5.77}$     & $-$
        \\%
        
        \hhline{|-|-||--|--|---|--|}%
      \end{tabular}%

%% file: Tables/AugmentedFlowNetC_results.tex
\resizebox{\textwidth}{!}{%
      \begin{tabular}{|c|l||c|c||c|c|}%
              \hhline{|-|-||--T--|}%
               & 
               & \multicolumn{2}{c||}{Animation Domain}%
               & \multicolumn{2}{c|}{Driving Domain}%
               \\%
               
               & 
               & \multicolumn{2}{c||}{Sintel}%
               & \multicolumn{2}{c|}{KITTI}%
               \\%
               
               & Method
               & \textit{clean}%
               & \textit{final}%
               & 2012%
               & 2015%
               \\%
               
            \hhline{|=|=||==#==|}%

        &
        Baseline
        & $3.07$  & $4.46$
        & $6.66$  & $12.47$
        \\%
        \hhline{|=|=||==#==|}%
        \multirow{8}{*}{\rotatebox[origin=c]{90}{Dom.}} &
        AugmentedFlowNetD-FlowNet2
        & $\textbf{2.79}$  & $\textbf{4.05}$
        & $4.26$  & $\pz{9.60}$
        \\%

        &
        AugmentedFlowNetD-FlowFields
        & $2.97$  & $4.18$
        & $\textbf{3.62}$  & $\pz\textbf{8.01}$
        \\%
        
        &
        AugmentedFlowNetD-EpicFlow
        & $3.09$  & $4.21$
        & $3.70$  & $\pz{8.11}$
        \\%
        
        &
        AugmentedFlowNetD-DeepFlow
        & $3.34$  & $4.45$
        & $4.90$  & $12.11$
        \\%
        
        &
        AugmentedFlowNetD-LDOF
        & $3.66$  & $4.69$
        & $7.93$  & $14.93$
        \\%
        
        &
        AugmentedFlowNetD-Rand. Mix
        & $3.07$  & $4.20$
        & $4.14$  & $\pz{9.61}$
        \\%
        \hhline{|~|-||--|--|}%
        &
        AugmentedFlowNetD-FusionNet (L1)
        & $2.80$  & $3.97$
        & $3.69$  & $\pz{8.17}$
        \\%
        
        &
         \cellcolor{gray!15}AugmentedFlowNetD-FusionNet (Hinge)
        &  \cellcolor{gray!15}$\textbf{2.75}$  &  \cellcolor{gray!15}$\textbf{3.97}$
        &  \cellcolor{gray!15}$\textbf{3.52}$  &  \cellcolor{gray!15}$\pz\textbf{7.65}$
        \\%
        
        \hhline{|-|-||--|--|}%
        
        \multirow{2}{*}{\rotatebox[origin=c]{90}{Gen.}} &
        AugmentedFlowNetG-FusionNet (L1)
        & $2.84$  & $4.06$
        & $3.91$  & $\pz{8.34}$
        \\%
        
        &
         \cellcolor{gray!15}AugmentedFlowNetG-FusionNet (Hinge)
        &  \cellcolor{gray!15}$\textbf{2.78}$  &  \cellcolor{gray!15}$\textbf{4.02}$
        &  \cellcolor{gray!15}$\textbf{3.77}$  &  \cellcolor{gray!15}$\pz\textbf{8.01}$
        \\%
        
        \hhline{|-|-||--|--|}%
        
      \end{tabular}%
    }%

%% file: Tables/AugmentedFlowNetStacks_results.tex
      \begin{tabular}{|c|l||c|c||c|c|}%
              \hhline{|-|-||--T--|}%
               & 
               & \multicolumn{2}{c||}{Animation Domain}%
               & \multicolumn{2}{c|}{Driving Domain}%
               \\%
               
               & 
               & \multicolumn{2}{c||}{Sintel}%
               & \multicolumn{2}{c|}{KITTI}%
               \\%
               
               & Method
               & \textit{clean}%
               & \textit{final}%
               & 2012%
               & 2015%
               \\%
               
            \hhline{|=|=||==#==|}%

        &
        Baseline
        & $3.07$  & $4.46$
        & $6.66$  & $12.47$
        \\%
        
        \hhline{|=|=||==#==|}%
        
        \multirow{5}{*}{\rotatebox[origin=c]{90}{Dom.}} &
        UnFlow-C-CityScapes~\cite{unflow}
        & $-$  & $-$
        & $5.08$  & $10.78$
        \\%

        &
        UnFlow-C-ours
        & $4.22$  & $5.12$
        & $5.01$  & $\pz{11.07}$
        \\%
        
        &
        UnFlow-C-KITTIraw~\cite{unflow}
        & $-$  & $-$
        & $3.78$  & $\pz{8.80}$
        \\%
        
        &
        UnFlow-CS~\cite{unflow}
        & $-$  & $-$
        & $3.30$  & $\pz{8.14}$
        \\%
        
        &
        UnFlow-CSS~\cite{unflow}
        & $-$  & $-$
        & $\textbf{3.29}$  & $\pz\textbf{8.10}$
        \\%
        
        \hhline{|=|=||==#==|}%
        
        \multirow{3}{*}{\rotatebox[origin=c]{90}{Dom.}} &
        AugmentedFlowNetD-C
        & $2.75$  & $3.97$
        & $3.52$  & $\pz{7.65}$
        \\%
        
        &
        AugmentedFlowNetD-CS
        & $2.20$  & $3.45$
        & $2.35$  & $\pz{5.84}$
        \\%
        
        &
           \cellcolor{gray!15}AugmentedFlowNetD-CSS
        &  \cellcolor{gray!15}$\textbf{2.09}$  &  \cellcolor{gray!15}$\textbf{3.34}$
        &  \cellcolor{gray!15}$\textbf{2.05}$  &  \cellcolor{gray!15}$\pz\textbf{5.35}$
        \\%
        
        \hhline{|=|=||==#==|}%
        
        \multirow{3}{*}{\rotatebox[origin=c]{90}{Gen.}} &
        AugmentedFlowNetG-C
        & $2.78$  & $4.02$
        & $3.77$  & $\pz{8.01}$
        \\%
        
        &
        AugmentedFlowNetG-CS
        & $2.21$  & $3.49$
        & $2.44$  & $\pz{5.90}$
        \\%
        
        &
           \cellcolor{gray!15}AugmentedFlowNetG-CSS
        &  \cellcolor{gray!15}$\textbf{2.10}$  &  \cellcolor{gray!15}$\textbf{3.38}$
        &  \cellcolor{gray!15}$\textbf{2.17}$  &  \cellcolor{gray!15}$\pz\textbf{5.18}$
        \\%
        
        \hhline{|-|-||--|--|}%
        
      \end{tabular}%

%% file: Tables/Benchmark_results.tex
\resizebox{\textwidth}{!}{%
      \begin{tabular}{|c|l||cc|cc||ccc|cc|}%
              \hhline{|-|-||--|--||---T--|}%
               & 
               & \multicolumn{4}{c||}{Animation Domain}%
               & \multicolumn{5}{c|}{Driving Domain}%
               \\%
               
               & Method
               & \multicolumn{2}{c|}{Sintel \textit{clean}}%
               & \multicolumn{2}{c||}{Sintel \textit{final}}%
               & \multicolumn{3}{c|}{KITTI 2012}%
               & \multicolumn{2}{c|}{KITTI 2015}%
               \\%
               
               &
               & \multicolumn{2}{c|}{AEE}%
               & \multicolumn{2}{c||}{AEE}%
               & \multicolumn{2}{c}{AEE}%
               & F1-noc
               & AEE
               & Fl-all
               \\%
               
               &
               & \textit{train} & \textit{test}%
               & \textit{train} & \textit{test}%
               & \textit{train} & \textit{test} & \textit{test}%
               & \textit{train} & \textit{test}%
               \\%
               
        \hhline{|=|=||==|==||===|==|}%

        \multirow{4}{*}{\rotatebox[origin=c]{90}{Un/Semi}} &
        DSTFlow\cite{dstflow}%
        & $6.93$  & $5.20$
        & $7.82$  & $5.92$
        & $10.43$ & $12.4$ & $-$
        & $16.79$ & $39.00\%$
        \\%

        &
        GAN-OpticalFlow\cite{NIPS2017_6639}%
        & $3.30$ & $\phantom{^\dagger}6.27^\dagger$
        & $4.68$ & $\phantom{^\dagger}7.31^\dagger$
        & $\pz{7.16}$ & $\pz{6.8}$ & $-$
        & $16.02$ & $31.01\%$
        \\%
        
        &
        Hybrid-OpticalFlow-NextFrame\cite{SZB17}%
        & $-$ & $-$
        & $-$ & $-$
        & $\pz{5.31}$ & $\pz{9.2}$ & $\pz{39.12\%}$
        & $10.19$ & $-$
        \\%
        
        &
        UnFlow-CSS\cite{unflow}%
        & $-$ & $-$
        & $7.91$ & $10.22$
        & $\pz{3.29}$ & $\pz{\phantom{^\dagger}1.7^\dagger}$ & $\pz{\phantom{^\dagger}4.28\%^\dagger}$
        & $\pz{8.10}$ & $\phantom{^\dagger}11.11\%^\dagger$
        \\%
        
        \hhline{|-|-||--|--||---|--|}%
        
        \multirow{2}{*}{\rotatebox[origin=c]{90}{Sup}} &
        FlowNet2\cite{flownet2}%
        & $2.02$  & $3.96$
        & $3.62$  & $\phantom{^\dagger}5.74^\dagger$
        & $\pz{3.55}$  & $\pz{\phantom{^\dagger}1.8^\dagger}$ & $-$
        & $\pz{8.94}$  & $\phantom{^\dagger}11.48\%^\dagger$
        \\%

        &
        PWC-Net\cite{PWCnet}%
        & $2.55$ & $\phantom{^\dagger}3.86^\dagger$
        & $3.93$ & $\phantom{^\dagger}\textbf{5.04}^\dagger$
        & $\pz{4.14}$ & $\pz{\phantom{^\dagger}1.7^\dagger}$ & $\pz{\phantom{^\dagger}4.22\%^\dagger}$
        & $10.35$ & $\pz{\phantom{^\dagger}9.60\%^\dagger}$
        \\%
        
        \hhline{|-|-||--|--||---|--|}%
        
        \multirow{2}{*}{\rotatebox[origin=c]{90}{}} &
        DCFlow\cite{dcflow}%
        & $-$  & $3.54$
        & $-$  & $5.12$
        & $-$ & $-$ & $-$
        & $-$ & $14.86\%$
        \\%

        &
        MR-Flow\cite{Wulff:CVPR:2017}%
        & $1.83$ & $\textbf{2.53}$
        & $3.59$ & $5.38$
        & $-$ & $-$ & $-$
        & $-$ & $12.19\%$
        \\%
        
        \hhline{|-|-||--|--||---|--|}%
        
        \multirow{3}{*}{\rotatebox[origin=c]{90}{Our}} &
        AugmentedFlowNetD-CSS
        & $2.09$  & $\phantom{^\dagger}4.22^\dagger$
        & $3.34$  & $\phantom{^\dagger}5.63^\dagger$
        & $\pz\textbf{2.05}$ & $\phantom{^\dagger}\pz\textbf{1.5}^\dagger$ & $\pz{\phantom{^\dagger}\textbf{3.97\%}^\dagger}$
        & $\pz{5.35}$ & $\pz{\phantom{^\dagger}\textbf{8.57\%}^\dagger}$
        \\%

        &
        AugmentedFlowNetG-CSS%
        & $2.10$ & $3.69$
        & $3.38$ & $5.20$
        & $\pz{2.17}$ & $\pz{2.7}$ & $\pz{7.04\%}$
        & $\pz\textbf{5.18}$ & $20.09\%$
        \\%
        
        &
        FusionNet-Hinge%
        & $\textbf{1.58}$ & $3.20$
        & $\textbf{3.18}$ & $5.50$
        & $\pz{2.97}$ & $\pz{3.6}$ & $\pz{5.50\%}$
        & $\pz{8.18}$ & $21.44\%$
        \\%
        
        \hhline{|-|-||--|--||---|--|}%
      \end{tabular}%
    }%

%% file: Tables/MotionSeg_results.tex
    \begin{tabular}{|l||c|c|}%
              \hhline{|-||--|}%
               
               
               Method
               & F1 Measure%
               & Extracted Objects%
               \\%
               
            \hhline{|=||==|}%

        FlowNetC (baseline)
        & $60.52\%$  & $10/69$
        \\%
        
        \hhline{|-||--|}%
        
        AugmentedFlowNetD-C
        & $65.29\%$  & $13/69$
        \\%
        
        \hhline{|-||--|}%
        
        FlowNet2~\cite{flownet2} (stacked baseline)
        & $76.72\%$  & $26/69$
        \\%
        
        \hhline{|-||--|}%
        
        AugmentedFlowNetD-CS
        & $\textbf{77.09}\%$  & $\textbf{28/69}$
        \\%
        
        \hhline{|-||--|}%

      \end{tabular}%

%% file: eccv2016submission.bbl
\begin{thebibliography}{10}

\bibitem{flownet2}
Ilg, E., Mayer, N., Saikia, T., Keuper, M., Dosovitskiy, A., Brox, T.:
\newblock Flownet 2.0: Evolution of optical flow estimation with deep networks.
\newblock In: IEEE Conference on Computer Vision and Pattern Recognition
  (CVPR). (2017)

\bibitem{PWCnet}
Sun, D., Yang, X., Liu, M., Kautz, J.:
\newblock {PWC-Net: CNNs for Optical Flow Using Pyramid, Warping, and Cost
  Volume}.
\newblock In: IEEE Conference on Computer Vision and Pattern Recognition
  (CVPR). (2018)

\bibitem{schunck}
Horn, B.K.P., Schunck, B.G.:
\newblock Determining optical flow.
\newblock In: Artificial Intelligence (AI). (1981)

\bibitem{meminperez}
Mémin, E., Pérez, P.:
\newblock A multigrid approach for hierarchical motion estimation.
\newblock In: IEEE Int. Conference on Computer Vision (ICCV). (1998)

\bibitem{ahmadi}
Ahmadi, A., Patras, I.:
\newblock Unsupervised convolutional neural networks for motion estimation.
\newblock In: International Conference on Image Processing (ICIP). (2016)

\bibitem{unflow}
Meister, S., Hur, J., Roth, S.:
\newblock {UnFlow}: Unsupervised learning of optical flow with a bidirectional
  census loss.
\newblock In: Conference on Artificial Intelligence (AAAI). (2018)

\bibitem{SZB17}
Sedaghat, N., Zolfaghari, M., Brox, T.:
\newblock Hybrid learning of optical flow and next frame prediction to boost
  optical flow in the wild.
\newblock Technical report (2017)

\bibitem{guided_flow_17}
Zhu, Y., Lan, Z., Newsam, S., Hauptmann, A.G.:
\newblock {Guided Optical Flow Learning}.
\newblock arXiv preprint arXiv:1702.022952 (2017)

\bibitem{lucaskanade}
Lucas, B.D., Kanade, T.:
\newblock An iterative image registration technique with an application to
  stereo vision.
\newblock In: Imaging Understanding Workshop. (1981)

\bibitem{Bro04a}
Brox, T., Bruhn, A., Papenberg, N., Weickert, J.:
\newblock High accuracy optical flow estimation based on a theory for warping.
\newblock In: European Conference on Computer Vision (ECCV). (2004)

\bibitem{pocktvl1}
Zach, C., Pock, T., Bischof, H.:
\newblock {A duality based approach for realtime TV-L 1 optical flow}.
\newblock In: Joint Pattern Recognition Symposium. (2007)

\bibitem{ldof}
Brox, T., Malik, J.:
\newblock Large displacement optical flow: descriptor matching in variational
  motion estimation.
\newblock (2011)

\bibitem{deepmatching}
Weinzaepfel, P., Revaud, J., Harchaoui, Z., Schmid, C.:
\newblock {DeepFlow: Large displacement optical flow with deep matching}.
\newblock In: {IEEE Intenational Conference on Computer Vision (ICCV)}. (2013)

\bibitem{epicflow}
Revaud, J., Weinzaepfel, P., Harchaoui, Z., Schmid, C.:
\newblock {EpicFlow: Edge-Preserving Interpolation of Correspondences for
  Optical Flow}.
\newblock In: IEEE Conference on Computer Vision and Pattern Recognition
  (CVPR). (2015)

\bibitem{flowfields}
Bailer, C., Taetz, B., Stricker, D.:
\newblock Flow fields: Dense correspondence fields for highly accurate large
  displacement optical flow estimation.
\newblock In: IEEE Int. Conference on Computer Vision (ICCV). (2015)

\bibitem{dcflow}
Xu, J., Ranftl, R., Koltun, V.:
\newblock {Accurate Optical Flow via Direct Cost Volume Processing}.
\newblock In: IEEE Conference on Computer Vision and Pattern Recognition
  (CVPR). (2017)

\bibitem{Wulff:CVPR:2017}
Wulff, J., Sevilla-Lara, L., Black, M.J.:
\newblock Optical flow in mostly rigid scenes.
\newblock In: IEEE Conference on Computer Vision and Pattern Recognition
  (CVPR). (2017)

\bibitem{flownet}
Dosovitskiy, A., Fischer, P., Ilg, E., H{\"a}usser, P., Haz{\i}rba{\c{s}}, C.,
  Golkov, V., v.d. Smagt, P., Cremers, D., Brox, T.:
\newblock Flownet: Learning optical flow with convolutional networks.
\newblock In: IEEE Int. Conference on Computer Vision (ICCV). (2015)

\bibitem{dispnet}
Mayer, N., Ilg, E., H\"{a}usser, P., Fischer, P., Cremers, D., Dosovitskiy, A.,
  Brox, T.:
\newblock A large dataset to train convolutional networks for disparity,
  optical flow, and scene flow estimation.
\newblock In: IEEE Conference on Computer Vision and Pattern Recognition
  (CVPR). (2016)

\bibitem{spynet}
Ranjan, A., Black, M.:
\newblock Optical flow estimation using a spatial pyramid network.
\newblock In: IEEE Conference on Computer Vision and Pattern Recognition
  (CVPR). (2017)

\bibitem{FlowFieldsCNN}
Bailer, C., Varanasi, K., Stricker, D.:
\newblock {CNN-based Patch Matching for Optical Flow with Thresholded Hinge
  Embedding Loss}.
\newblock In: IEEE Conference on Computer Vision and Pattern Recognition
  (CVPR). (2017)

\bibitem{PatchBatch}
Gadot, D., Wolf, L.:
\newblock {PatchBatch: a Batch Augmented Loss for Optical Flow}.
\newblock In: IEEE Conference on Computer Vision and Pattern Recognition
  (CVPR). (2016)

\bibitem{DeepDiscreteFlow}
Güney, F., Geiger, A.:
\newblock {Deep Discrete Flow}.
\newblock In: Asian Conference on Computer Vision (ACCV). (2016)

\bibitem{backtobasics}
Yu, J.J., Harley, A.W., Derpanis, K.G.:
\newblock Back to basics: Unsupervised learning of optical flow via brightness
  constancy and motion smoothness.
\newblock In: European Conference on Computer Vision (ECCV). (2016)

\bibitem{dstflow}
Ren, Z., Yan, J., Ni, B., Liu, B., Yang, X., Zha, H.:
\newblock Unsupervised deep learning for optical flow estimation.
\newblock In: Conference on Artificial Intelligence (AAAI). (2017)

\bibitem{NIPS2017_6639}
Lai, W.S., Huang, J.B., Yang, M.H.:
\newblock Semi-supervised learning for optical flow with generative adversarial
  networks.
\newblock In: Advances in Neural Information Processing Systems 30.
\newblock (2017)

\bibitem{lempitskyCVPR08}
Lempitsky, V., Roth, S., Rother, C.:
\newblock {FusionFlow: Discrete-Continuous Optimization for Optical Flow
  Estimation}.
\newblock In: IEEE Conference on Computer Vision and Pattern Recognition
  (CVPR). (2008)

\bibitem{mdpflow}
Xu, L., Jia, J., Matsushita, Y.:
\newblock Motion detail preserving optical flow estimation.
\newblock In: IEEE Transactions on Pattern Analysis and Machine Intelligence
  (TPAMI). (2012)

\bibitem{NIPS2003_2366}
Schultz, M., Joachims, T.:
\newblock Learning a distance metric from relative comparisons.
\newblock In: Advances in Neural Information Processing Systems 16.
\newblock (2004)

\bibitem{weinberger2009distance}
Weinberger, K.Q., Saul, L.K.:
\newblock Distance metric learning for large margin nearest neighbor
  classification.
\newblock JMLR (2009)

\bibitem{wang01}
Wang, J., Song, Y., Leung, T., Rosenberg, C., Wang, J., Philbin, J., Chen, B.,
  Wu, Y.:
\newblock Learning fine-grained image similarity with deep ranking.
\newblock In: IEEE Conference on Computer Vision and Pattern Recognition
  (CVPR). (2014)

\bibitem{tripletnet}
Hoffer, E., Ailon, N.:
\newblock Deep metric learning using triplet network.
\newblock In: ICLR (Workshop). (2015)

\bibitem{fastnet}
Schroff, F., Kalenichenko, D., Philbin, J.:
\newblock Facenet: A unified embedding for face recognition and clustering.
\newblock In: IEEE Conference on Computer Vision and Pattern Recognition
  (CVPR). (2015)

\bibitem{wohlhart01}
Wohlhart, P., Lepetit, V.:
\newblock Learning descriptors for object recognition and 3d pose estimation.
\newblock In: IEEE Conference on Computer Vision and Pattern Recognition
  (CVPR). (2015)

\bibitem{multihinge}
Moore, R.C., DeNero, J.:
\newblock L1 and l2 regularization for multiclass hinge loss models.
\newblock In: Symposium on Machine Learning in Speech and Natural Language
  Processing. (2011)

\bibitem{multisvm}
Do\u{g}an, {\"U}., Glasmachers, T., Igel, C.:
\newblock A unified view on multi-class support vector classification.
\newblock Journal of Machine Learning Research (2016)

\bibitem{blender}
Project, B.:
\newblock Open movies (agent, caminandes, cosmos, sintel and big buck bunny).
\newblock \url{https://www.blender.org/about/projects} (2017)

\bibitem{Butler:ECCV:2012}
Butler, D.J., Wulff, J., Stanley, G.B., Black, M.J.:
\newblock A naturalistic open source movie for optical flow evaluation.
\newblock In: European Conference on Computer Vision (ECCV). (2012)

\bibitem{MIFDB16}
N.Mayer, E.Ilg, P.H{\"a}usser, P.Fischer, D.Cremers, A.Dosovitskiy, T.Brox:
\newblock A large dataset to train convolutional networks for disparity,
  optical flow, and scene flow estimation.
\newblock In: IEEE Conference on Computer Vision and Pattern Recognition
  (CVPR). (2016)

\bibitem{Cordts2016Cityscapes}
Cordts, M., Omran, M., Ramos, S., Rehfeld, T., Enzweiler, M., Benenson, R.,
  Franke, U., Roth, S., Schiele, B.:
\newblock The cityscapes dataset for semantic urban scene understanding.
\newblock In: IEEE Conference on Computer Vision and Pattern Recognition
  (CVPR). (2016)

\bibitem{Geiger2012CVPR}
Geiger, A., Lenz, P., Urtasun, R.:
\newblock {Are we ready for Autonomous Driving? The KITTI Vision Benchmark
  Suite}.
\newblock In: IEEE Conference on Computer Vision and Pattern Recognition
  (CVPR). (2012)

\bibitem{Menze2015CVPR}
Menze, M., Geiger, A.:
\newblock Object scene flow for autonomous vehicles.
\newblock In: IEEE Conference on Computer Vision and Pattern Recognition
  (CVPR). (2015)

\bibitem{udg}
Mahmood, M.H., Díez, Y., Salvi, J., Lladó, X.:
\newblock A collection of challenging motion segmentation benchmark datasets.
\newblock (2017)

\bibitem{OB14b}
Ochs, P., Malik, J., Brox, T.:
\newblock Segmentation of moving objects by long term video analysis.
\newblock IEEE Transactions on Pattern Analysis and Machine Intelligence
  (TPAMI) (2014)

\bibitem{KB15b}
Keuper, M., Andres, B., Brox, T.:
\newblock Motion trajectory segmentation via minimum cost multicuts.
\newblock In: IEEE Int. Conference on Computer Vision (ICCV). (2015)

\end{thebibliography}
